\documentclass[twoside]{article}

 \usepackage[accepted]{aistats2020_arxiv}
 \newcommand{\version}[2]{#1} %

\usepackage[round]{natbib}

\usepackage[utf8]{inputenc} %
\usepackage[T1]{fontenc}    %
\usepackage{url}            %
\usepackage{booktabs}       %
\usepackage{amsfonts}       %
\usepackage{nicefrac}       %
\usepackage{microtype}      %
\usepackage{amsmath}
\usepackage{algorithm}
\usepackage{algpseudocode}
\usepackage{eqparbox}

\usepackage{graphicx}
\usepackage{subcaption}
\usepackage{bm}
\usepackage{bbm}
\usepackage{color}
\usepackage[shortlabels]{enumitem}
\usepackage{tikz}
\usepackage{pgfplots}
\usetikzlibrary{pgfplots.groupplots}
\usetikzlibrary{shapes}
\usepackage{tabularx}
\usepackage{todonotes}
\usepackage{multirow}
\usepackage{glossaries}
\usepackage{todonotes}
\usepackage{wrapfig}
\usepackage{lipsum}

\glsdisablehyper

\definecolor{tableauC0}{rgb}{0.122, 0.467, 0.706}
\definecolor{tableauC1}{rgb}{1.000, 0.500, 0.055}
\definecolor{tableauC2}{rgb}{0.172, 0.627, 0.172}
\definecolor{tableauC3}{rgb}{0.839, 0.153, 0.157}
\definecolor{tableauC4}{rgb}{0.578, 0.402, 0.738}
\definecolor{tableauC5}{rgb}{0.547, 0.336, 0.293}
\definecolor{tableauC6}{rgb}{0.887, 0.465, 0.759}
\definecolor{tableauC7}{rgb}{0.500, 0.500, 0.500}
\definecolor{tableauC8}{rgb}{0.734, 0.738, 0.133}

\tikzset{cross/.style={cross out, draw=black, minimum size=2*(#1-\pgflinewidth), inner sep=0pt, outer sep=0pt}, cross/.default={1pt}}

\newcommand\norm[1]{\left\lVert#1\right\rVert}
\newcommand{\lfalgref}[1]{Alg.\,\ref{#1}}
\newcommand{\lfeqref}[1]{Eq.\,\eqref{#1}}
\newcommand{\lffigref}[1]{Fig.\,\ref{#1}}
\newcommand{\lfsecref}[1]{Sec.\,\ref{#1}}

\newcommand{\indicator}[1]{\mathbbm{1}_{\{#1\}}}
\newcolumntype{Y}{>{\centering\arraybackslash}X}
\newcommand{\datan}{\mathcal{D}_n}
\newcommand{\figurefontsize}{\scriptsize}

\newacronym{acr:bo}{BO}{Bayesian optimization}
\newacronym{acr:ei}{EI}{expected improvement}
\newacronym{acr:ep}{EP}{expectation propagation}
\newacronym{acr:es}{ES}{entropy search}
\newacronym{acr:gp}{GP}{Gaussian process}
\newacronym{acr:ir}{IR}{inference regret}
\newacronym{acr:kde}{KDE}{kernel density estimate}
\newacronym{acr:mes}{MES}{max-value entropy search}
\newacronym{acr:nes}{NES}{Noisy-Input Entropy Search}
\newacronym{acr:pes}{PES}{predictive entropy search}
\newacronym{acr:pi}{PI}{probability of improvement}
\newacronym{acr:rs}{RS}{rejection sampling}
\newacronym{acr:se}{SE}{squared exponential}
\newacronym{acr:ssgp}{SSGP}{sparse spectrum Gaussian process}
\newacronym{acr:ucb}{UCB}{upper confidence bound}

\begin{document}

\runningtitle{Noisy-Input Entropy Search for Efficient Robust Bayesian Optimization}

\runningauthor{Fr\"ohlich, Klenske, Vinogradska, Daniel, Zeilinger}

\twocolumn[

\aistatstitle{Noisy-Input Entropy Search\\for Efficient Robust Bayesian Optimization}
\aistatsauthor{Lukas P. Fr\"ohlich\textsuperscript{\textnormal{1,2}} \And Edgar D. Klenske\textsuperscript{\textnormal{1}} \And Julia Vinogradska\textsuperscript{\textnormal{1}} \AND Christian Daniel\textsuperscript{\textnormal{1}} \And Melanie N. Zeilinger\textsuperscript{\textnormal{2}} }
\aistatsaddress{\textsuperscript{1}Bosch Center for Artificial Intelligence \\ Renningen, Germany \And \textsuperscript{2}ETH Z\"urich \\ Z\"urich, Switzerland}
]

\begin{abstract}
We consider the problem of robust optimization within the well-established \gls{acr:bo} framework.
While \gls{acr:bo} is intrinsically robust to noisy evaluations of the objective function, standard approaches do not consider the case of uncertainty about the input parameters.
In this paper, we propose \textit{\gls{acr:nes}}, a novel information-theoretic acquisition function that is designed to find robust optima for problems with both input and measurement noise.
\gls{acr:nes} is based on the key insight that the robust objective in many cases can be modeled as a Gaussian process, however, it cannot be observed directly.
We evaluate \gls{acr:nes} on several benchmark problems from the optimization literature and from engineering.
The results show that \gls{acr:nes} reliably finds robust optima, outperforming existing methods from the literature on all benchmarks.
\end{abstract}

\section{Introduction}
\acrfull{acr:bo} is a well-established technique for optimization of black-box functions with applications in a wide range of domains \citep{Brochu2010TutorialBO,Shahriari2016BayesianOptimization}.
The two key benefits of \gls{acr:bo} are its sample-efficiency and its intrinsic robustness to noisy function evaluations, rendering it particularly powerful when function evaluations are either time consuming or costly, e.g., for drug design or (robot) controller tuning \citep{Calandra2016GaitOptimization, Cully2015nature,Griffiths2017AutomaticChemicalDesignBO}.
The sample-efficiency of \gls{acr:bo} stems from two key ingredients: (i) a Bayesian surrogate model that approximates the objective function based on previous evaluations, e.g., \gls{acr:gp} regression, and (ii) an acquisition function that defines the next evaluation point based on the surrogate model.
Several acquisition functions have been proposed that heuristically trade off between exploration and exploitation \citep{Kushner1964PI, Mockus1975ExpectedImprovement, Cox1992UpperConfidenceBound}.
More recently, the family of entropy search acquisition functions has been introduced.
These acquisition functions use an information-theoretic approach and choose the next evaluation point to maximize information about the global optimum.
At the cost of computational complexity, entropy-based methods are generally more sample-efficient than other acquisition functions \citep{Hennig2012EntropySearch, HernandezLobato2014PredictiveEntropySearch, Wang2017MaxValueEntropySearch}.

In addition to sample efficiency, robustness with respect to model uncertainties or perturbations on the input is critical in many applications (see, e.g., \citep{Beyer2007RobustOptimizationSurvey} for a survey).
Examples are numerous in fields such as control \citep{Bacsar2008HInfinityControl}, design engineering \citep{Chen1996RobustDesignEngineering} and operations research \citep{Adida2006RobustOptimizationLogistics}.
In its standard formulation, \gls{acr:bo} is intrinsically robust with respect to noisy function evaluations, however, it leads to sub-optimal solutions in the presence of perturbations on the input.
While robust optimization has been considered in the context of \gls{acr:bo} before, previous work is based on heuristic acquisition functions.
To the best of our knowledge, entropy-based acquisition functions for robust optimization problems that fully leverage the potential of \gls{acr:bo} have not been addressed to date.

\paragraph{Contributions}
In this paper, we introduce the first entropy-based acquisition function that addresses the problem of robust optimization.
We consider a probabilistic formulation of robustness where the parameters found during optimization are randomly perturbed at the implementation stage, i.e., broad optima are preferable over narrow ones.
Due to their sample-efficiency, we build on entropy-based acquisition functions and propose to choose the next evaluation point in order to maximize the information about the robust optimum.
Our method is based on the key insight that the robust objective can be modeled with a \gls{acr:gp} just as in the standard \gls{acr:bo} setting.
However, the robust objective is not directly observable, but needs to be constructed from (noisy) evaluations of the original function without perturbations on the input.
We base our framework on the recently proposed \gls{acr:mes} \citep{Wang2017MaxValueEntropySearch}, due to the low computational demand.
The resulting formulation requires knowledge of the \gls{acr:gp}'s predictive distribution conditioned on the robust maximum value, which is an analytically intractable distribution.
We propose two methods to approximate this distribution
(i) based on rejection sampling, which in the limit of infinitely many samples is exact but computationally expensive, and 
(ii) based on \gls{acr:ep} \citep{Minka2001ExpectationPropagation}, which is computationally more efficient.
We evaluate the proposed acquisition function on a wide range of benchmark problems and compare against related approaches from the literature.
Moreover, we apply the proposed method to a simulated aerospace task to demonstrate the importance of robust black-box optimization in practice.

\paragraph{Related Work}
Closely related to our method are the approaches presented by \citet{Nogueira2016unscentedBO} and \citet{Beland2017UncertaintyBONipsWorkshop}, both of which consider the same probabilistic robust objective as considered in this paper (see \lfeqref{eq:robust_optimization}).
\citet{Nogueira2016unscentedBO} proposed to use the expectation of the \gls{acr:ei} acquisition function with respect to the input noise.
The expectation is approximated using the unscented transformation \citep{Julier2004UnscentedFiltering}, which is computationally efficient, but the approximation accuracy strongly depends on the choice of hyperparameters.
In the paper by \citet{Beland2017UncertaintyBONipsWorkshop}, the robust objective is also modeled as a \gls{acr:gp}, however, it is implicitly assumed that the robust objective can be observed directly.
In contrast to the two aforementioned methods, our method uses an information-theoretic approach.
We compare our method to both \citet{Nogueira2016unscentedBO} and \citet{Beland2017UncertaintyBONipsWorkshop}.

Besides random perturbations on the optimization parameters, other robust optimization settings have been investigated in the context of \gls{acr:bo}.
In recent work, \citet{Bogunovic2018AdversariallyBO} consider the worst-case perturbation within a given set (or minimax setting) instead of random perturbations.
Moreover, the authors provide rigorous convergence guarantees for their acquisition function, based on the results from \citet{Srinivas2010UpperConfidenceBound}.
\citet{Chen2017RobustNonConvexObjectives} consider a finite set of non-convex objective functions and seek the maximizer that is robust with respect to the choice of objective function from the given set.
In the setting considered by \citet{Martinez2018OutlierBO}, some evaluations are corrupted such that their value is perturbed much stronger than the observation noise, thus biasing the surrogate model.
However, this setting does not extend to the case of perturbations on the input.
\citet{Groot2010BayesianMonteCarloBO,Tesch2011SnakeGaitBOChangingEnvironments,Toscano2018ExpensiveIntegralBO} assume that the objective function depends on two types of input parameters: the control parameters to be optimized and environmental parameters against which the maximizer should be robust.
This differs from our setting, in which we aim at finding an optimum that is robust with respect to the control parameters.

Similar to \citet{Nogueira2016unscentedBO,Beland2017UncertaintyBONipsWorkshop,Bogunovic2018AdversariallyBO}, we assume exact knowledge of the control parameters during the optimization and require robustness when deploying the optimal parameters.
In contrast, \citet{Oliveira2019UncertainInputBO} proposed a method that deals with uncertain inputs during the optimization process, however, their goal is to find the global optimum instead of the robust optimum.

\section{Preliminaries}
In this section, we briefly review \acrfull{acr:bo} and discuss how it relates to the robust optimization setting considered in this paper.
As the robust objective will be approximated with \gls{acr:gp} regression, we furthermore summarize how perturbations on the input parameters can be included in the posterior predictive distribution.

\paragraph{Bayesian Optimization}\label{sec:bayesian_optimization}
In \gls{acr:bo} we seek the maximizer of the unknown objective function \mbox{$f(\bm{x}): \mathcal{X} \rightarrow \mathbb{R}$} over a compact set $\mathcal{X} \subseteq \mathbb{R}^{d}$ despite only having access to noisy observations, $y_i = f(\bm{x}_i) + \epsilon$ with $\epsilon \sim \mathcal{N}(0, \sigma_\epsilon^2)$.
Furthermore, no gradient information is available and each evaluation of $f(\bm{x})$ takes a considerable amount of time or effort.
Thus, the goal is to find the maximum in as few evaluations as possible.
The core idea of \gls{acr:bo} is to model the unknown objective $f(\bm{x})$ with a Bayesian surrogate model based on past observations $\datan = \{(\bm{x}_i, y_i)\}_{i=1:n}$.
Common choices for the model are Bayesian neural networks \citep{Snoek2015DNGO} or \glspl{acr:gp} \citep{Rasmussen2006Book}. In this paper, we consider the latter.
Based on the surrogate model, the next query point is chosen by maximizing a so-called acquisition function $\alpha(\bm{x})$.

Acquisition functions quantify the exploration-exploitation trade-off between regions with large predicted values (exploitation) and regions of high uncertainty (exploration).
Entropy-based acquisition functions address this trade-off by minimizing the uncertainty of the belief about global optimum's location $p(\bm{x}^* | \datan)$ \citep{Hennig2012EntropySearch}.
The next evaluation point $\bm{x}_{n+1}$ is chosen to maximize the mutual information between the global optimum $\bm{x}^*$ and the next evaluation point, given by $I( (\bm{x}, y) ; \bm{x}^* | \datan)$.
Recently, \citet{Wang2017MaxValueEntropySearch} introduced the \gls{acr:mes} acquisition function, which considers the optimum's value $y^*$ instead of its location, i.e., $\alpha_{\text{MES}}(\bm{x}) = I( (\bm{x}, y) ; y^* | \datan)$.
This formulation significantly reduces the computational burden compared to its predecessors.

By design, \gls{acr:bo} is able to efficiently optimize non-convex black-box functions.
However, it is generally not able to find optima that are robust with respect to perturbations of the input parameters.

\paragraph{Robust Bayesian Optimization}\label{sec:robust_bayesian_optimization}
In this paper, we consider a probabilistic formulation of robustness, i.e., we assume that the optimization parameters are randomly perturbed at implementation stage.
In the presence of input noise, broad optima should be preferred over narrow ones.
Thus, instead of optimizing $f(\bm{x})$ directly, we aim at maximizing the \textit{robust objective},
\begin{align}\label{eq:robust_optimization}
g(\bm{x}) = \mathop{\mathbb{E}}_{\bm{\xi} \sim p(\bm{\xi})}\left[ f(\bm{x} + \bm{\xi}) \right] = \int f(\bm{x} + \bm{\xi}) p(\bm{\xi}) d \bm{\xi},
\end{align}
such that the robust optimizer is given by \mbox{$\bm{x}^* = \arg \max_{\bm{x} \in \mathcal{X}} g(\bm{x})$}.
The random perturbations acting on the input parameters $\bm{x}$ are characterized by the distribution $p(\bm{\xi})$.
To this end, we assume $p(\bm{\xi}) \sim \mathcal{N}(0, \bm{\Sigma}_x)$ with $\bm{\Sigma}_x = \operatorname{diag}[\sigma_{x, 1}^2, \dots, \sigma_{x, d}^2]$ and $\sigma^2_{x,i}$ to be known for all $i$.
Other choices are of course possible, e.g., $p(\bm{\xi})$ could be chosen as a uniform distribution.
Note that for vanishing input noise, $\sigma_{x,i} \rightarrow 0$ for all $i$, the distribution $p(\bm{\xi})$ converges to the Dirac delta distribution and we obtain the standard, non-robust optimization setting.

\paragraph{Gaussian Process Regression}\label{sec:gp_regression}
\acrfull{acr:gp} regression is a non-parametric method to model an unknown function $f(\bm{x}): \mathcal{X} \mapsto \mathbb{R}$ from data $\datan$ (see, e.g., \citep{Rasmussen2006Book}).
A \gls{acr:gp} defines a prior distribution over functions, such that any finite number of function values are normally distributed with mean $\mu_f(\bm{x})$ and covariance specified by the kernel function $k_f(\bm{x}, \bm{x}')$ for any $\bm{x}, \bm{x}' \in \mathcal{X}$ (w.l.o.g. we assume $\mu_f(\bm{x}) \equiv 0$).
Conditioning the prior distribution on observed data $\datan$ leads to the posterior predictive mean and variance,
\begin{align}
\begin{split}
m_f(\bm{x} | \datan) & = \bm{k}_f(\bm{x})^\top \bm{K}^{-1} \bm{y}, \\
v_f(\bm{x} | \datan) & = k_f(\bm{x}, \bm{x}) - \bm{k}_f(\bm{x})^\top \bm{K}^{-1} \bm{k}_f(\bm{x}),
\end{split}
\end{align}
at any $\bm{x} \in \mathcal{X}$ with $[\bm{k}_f(\bm{x})]_i = k_f(\bm{x}, \bm{x}_i)$, $[\bm{K}]_{ij} = k_f(\bm{x}_i, \bm{x}_j) + \delta_{ij}\sigma_\epsilon^2$, $[\bm{y}]_i = y_i$ and $\delta_{ij}$ denotes the Kronecker delta.

In the context of Bayesian optimization, \gls{acr:gp} regression is commonly used as a surrogate model for the objective $f(\bm{x})$.
Since the expectation is a linear operator and \glspl{acr:gp} are closed under linear operations \citep{Rasmussen2006Book}, the robust objective $g(\bm{x})$ can be modeled as a \gls{acr:gp} as well, based on noisy observations of $f(\bm{x})$.
The predictive distribution for the robust objective then becomes
\begin{align}\label{eq:predictive_distribution_g}
\begin{split}
m_g(\bm{x} | \datan) & = \bm{k}_{gf}(\bm{x})^\top \bm{K}^{-1} \bm{y}, \\
v_g(\bm{x} | \datan) & = k_g(\bm{x}, \bm{x}) - \bm{k}_{gf}(\bm{x})^\top \bm{K}^{-1} \bm{k}_{fg}(\bm{x}),
\end{split}
\end{align}
where the respective kernel functions are given by $k_g(\bm{x}, \bm{x}') = \iint k_f(\bm{x} + \bm{\xi}, \bm{x'} + \bm{\xi'}) p(\bm{\xi})p(\bm{\xi'}) d\bm{\xi}d\bm{\xi'}$ and $k_{gf}(\bm{x}, \bm{x'}) = \int k_f(\bm{x} + \bm{\xi}, \bm{x'}) p(\bm{\xi}) d\bm{\xi}$.
For the well-known squared exponential and Mat\'ern kernel functions, $k_{gf}$ and $k_g$ can be computed in closed-form for normally and uniformly distributed input noise $\bm{\xi}$  (see, e.g., \citep{dallaire2009learning}).

\section{Noisy-Input Entropy Search}
In this section, we elaborate on the main contribution of this paper.
We first present our robust acquisition function and give an overview of the challenges associated with the proposed approach.
The main challenge is that the robust formulation requires the \gls{acr:gp}'s predictive distribution conditioned on the robust maximum value, which is analytically intractable.
We propose two approximation schemes: The first is based on \acrfull{acr:rs}, which gives the exact result in the limit of infinitely many samples, but is computationally challenging.
The second approach is based on \acrfull{acr:ep} \citep{Minka2001ExpectationPropagation} and is computationally more efficient, albeit not unbiased.

As discussed in \lfsecref{sec:bayesian_optimization}, entropy-based acquisition functions quantify the information gain about the global optimum of $f(\bm{x})$.
Hence, the next evaluation point $\bm{x}_{n+1}$ is selected to be maximally informative about $\bm{x}^*$ (or $y^*$ for \gls{acr:mes}).
For robust optimization, we aim at finding the maximizer of the robust objective $g(\bm{x})$ instead.
We build on the work of \citet{Wang2017MaxValueEntropySearch} and consider the mutual information between $\bm{x}$ and the objective's maximum value.
Consequently, we maximize the information about the robust maximum value $g^* = \max_{\bm{x} \in \mathcal{X}} g(\bm{x})$ and propose the \textit{\acrfull{acr:nes}} acquisition function
\begin{align}\label{eq:nes_acquisition_function}
&\alpha_{\text{NES}}(\bm{x}) = I\Big( (\bm{x}, y) ; g^* | \datan \Big) \notag \\
 & =  H \big[ p( y(\bm{x}) | \datan) \big] - \mathop{\mathbb{E}}_{g^* | \datan} \Big[ H \big[ p( y(\bm{x}) | \datan, g^* ) \big] \Big],
\end{align}
where $I(\cdot ; \cdot | \cdot)$ denotes the conditional mutual information and $H[\cdot]$ the differential entropy.
Note how \gls{acr:nes} reasons about $g^*$ while only (noisily) observing $f(\bm{x})$ as opposed to the na\"ive approach of applying \acrfull{acr:mes} to the \gls{acr:gp} model of the robust objective, which assumes access to observations of $g(\bm{x})$.
The corresponding mutual information would be $I((\bm{x}, z) ; g^* | \datan)$, with the hypothetical observation model $z = g(\bm{x}) + \eta$ and $\eta \sim \mathcal{N}(0, \sigma_\eta^2)$.
This, however, is not possible as $g(\bm{x})$ cannot be observed directly.

The first term in \lfeqref{eq:nes_acquisition_function} corresponds to the entropy of the \gls{acr:gp}'s predictive posterior distribution.
For Gaussian distributions, the entropy can be computed analytically such that $ H [ p( y(\bm{x}) | \datan)] = 0.5 \log[2 \pi e ( v_f(\bm{x} | \datan) + \sigma^2_\epsilon )]$.
The second term in \lfeqref{eq:nes_acquisition_function} has no analytic solution and requires approximations for the following reasons:
(i) The expectation is with respect to the unknown distribution over $g^*$ and (ii) it is not obvious how conditioning on the robust maximum value $g^*$ influences the predictive distribution $p( y(\bm{x}) | \datan)$.
In what follows we will address these challenges.

\subsection{Approximating the Expectation Over Robust Maximum Values}\label{sec:monte_carlo_approximation_expectation_max_vals}
The belief over the robust maximum value $p(g^* | \datan)$ in \lfeqref{eq:nes_acquisition_function} cannot be computed in closed form.
In the standard \gls{acr:bo} setting, the corresponding expectation has been approximated via Monte Carlo sampling \citep{HernandezLobato2014PredictiveEntropySearch,Wang2017MaxValueEntropySearch}
We follow this approach and approximate the expectation over $p(g^* | \datan)$ as
\begin{multline}\label{eq:monte_carlo_approximation_expectation_max_vals}
\mathop{\mathbb{E}}_{g^* | \datan} \Big[ H \big[ p( y(\bm{x}) | \datan, g^* ) \big] \Big] 
\approx \\ \frac{1}{K} \sum_{g^*_k \in G^*} H \big[ p( y(\bm{x}) | \datan, g^*_k ) \big],
\end{multline}
where $G^*$ is a set of $K$ samples drawn from $p(g^* | \datan)$.
We generate samples $g^*_k \in G^*$ via a two-step process: (i) sample a function $\tilde{g}_k(\bm{x})$ from $p(g(\bm{x}) | \datan)$ and (ii) maximize it such that $g^*_k = \max_{\bm{x} \in \mathcal{X}}\tilde{g}_k(\bm{x})$.

For efficient function sampling from $p(g(\bm{x}) | \datan)$ and subsequent maximization, we employ the \gls{acr:ssgp} approximation \citep{LazaroGredilla2010SparseSpectrumGP}.
The advantage of \glspl{acr:ssgp} is that the sampled functions can be  efficiently optimized with a gradient-based optimizer.
In this case, we can sample functions from $p(f(\bm{x}) | \datan)$ that are of the form $\tilde{f}_k(\bm{x}) = \bm{a}^T \bm{\phi}_f(\bm{x})$, where
$\bm{\phi}_f(\bm{x}) \in \mathbb{R}^M$ is a vector of random feature functions.
The components of the feature vector $\bm{\phi}_f(\bm{x}) \in \mathbb{R}^M$ are given by $\phi_{f,i}(\bm{x}) = \cos (\bm{w}_i^T \bm{x} + b_i)$, with $b_i \sim \mathcal{U}(0, 2\pi)$ and  $\bm{w}_i \sim p(\bm{w}) \propto s(\bm{w})$ where $s(\bm{w})$ is the Fourier dual of the kernel function $k_f$.
The weight vector $\bm{a}$ is distributed according to $\mathcal{N}(\bm{A}^{-1} \bm{\Phi}_f^T \bm{y}, \sigma^2_\epsilon \bm{A}^{-1})$ with $\bm{A} = \bm{\Phi}_f^T \bm{\Phi}_f + \sigma^2_\epsilon \bm{I}$, $\bm{\Phi}_f^T = [\bm{\phi}_f(\bm{x}_1), \dots, \bm{\phi}_f(\bm{x}_n)]$, $\datan = \{(\bm{x}_i, y_i)\}_{i=1:n}$ and $\bm{y} = [y_1, \dots, y_n]$ (see, e.g., \citep{LazaroGredilla2010SparseSpectrumGP} or \citep{HernandezLobato2014PredictiveEntropySearch} for details).

We can now generate $\tilde{g}_k(\bm{x})$ from a function sample $\tilde{f}_k(\bm{x})$ by taking the expectation w.r.t. the input noise.
As each $\tilde{f}_k(\bm{x})$ is a linear combination of $M$ cosine functions, we can compute this expectation in closed form.
For normally distributed input noise, $\bm{\xi} \sim \mathcal{N}(0, \bm{\Sigma}_x)$ with $\bm{\Sigma}_x = \operatorname{diag}[\sigma_{x, 1}^2, \dots, \sigma_{x, d}^2]$, this operation reduces to a scaling of the feature functions,
\begin{align}\label{eq:filtered_ssgp_basis_function}
\phi_{g, i}(\bm{x}) & = \int \phi_{f, i}(\bm{x} + \bm{\xi}) p(\bm{\xi}) d\bm{\xi} \notag \\
& = \phi_{f, i}(\bm{x})  \exp\Big( -\frac{1}{2} \sum_{j=1}^{d} \bm{w}_{i,j}^2 \sigma_{x,j}^2 \Big). 
\end{align}
Thus, we can efficiently sample $g^*_k \sim p(g^* | \datan)$ exploiting the fact that $g^*_k = \max_{\bm{x} \in \mathcal{X}} \tilde{g}_k(\bm{x})$ and $\tilde{g}_k(\bm{x}) = \bm{a}^T \bm{\phi}_g(\bm{x})$.
We present a detailed derivation of \lfeqref{eq:filtered_ssgp_basis_function} \version{in the appendix (\lfsecref{app:fourier_transform_ssgp})}{} as well as a discussion on the number of samples needed for a sufficient approximation accuracy of \lfeqref{eq:monte_carlo_approximation_expectation_max_vals} \version{(\lfsecref{app:num_maxvalue_samples})}{in the supplementary material}.

\subsection{Approximating the Conditional Predictive Distribution}
In the previous section we discussed how to sample robust maximum values $g^*_k \sim p(g^* | \datan)$.
To evaluate the proposed acquisition function $\alpha_{\text{NES}}(\bm{x})$, we need to compute the entropy of the predictive distribution conditioned on a sampled robust maximum value, i.e., $H[p(y(\bm{x}) | \datan, g^*_k)]$.
Conditioning on $g^*_k$ imposes $g(\bm{x}) \leq g^*_k$, which renders the computation of $p(y(\bm{x}) | \datan, g^*_k)$ intractable.
In this section, we propose two approximation schemes (i) based on \gls{acr:rs} which is exact in the limit of infinite samples and (ii) a computationally more efficient approach based on \gls{acr:ep} \citep{Minka2001ExpectationPropagation}.

\subsubsection{Using Rejection Sampling}
\label{sec:rs_based_approximation}

\begin{algorithm}[t]
 \caption{Rejection sampling for $p(y(\bm{x}) | \datan, g^*_k)$}
 \label{alg:rejection_sampling_robust_max_value}
 \begin{algorithmic}[1]
  \State \textbf{Input:} \gls{acr:gp} posterior predictive distribution $p(y(\bm{x}) | \datan)$, robust maximum value sample $g^*_k$
  \State \textbf{Output:} Set $\tilde{Y}$ of $L$ accepted samples 
  \State $\tilde{Y} \gets \emptyset$
  \While{$ |\tilde{Y}| \leq L $ } 
  \State $\tilde{f}(\bm{x}) \sim p(y(\bm{x}) | \datan)$ \texttt{ // Generate sample}
  \State $\tilde{g}(\bm{x}) \gets \int \tilde{f}(\bm{x} + \bm{\xi}) p(\bm{\xi})d\bm{\xi}$ \texttt{ // Robust sample}
  \If{$\max_{\bm{x}} \tilde{g}(\bm{x}) \leq g^*_k$}
  \State $\tilde{Y} \gets \tilde{Y} \cup \{\tilde{f}(\bm{x}) + \epsilon\}$ \texttt{// Store sample}
  \EndIf
  \EndWhile \\
  \Return $\tilde{Y}$
 \end{algorithmic}
\end{algorithm}

\begin{figure*}[h!]
 \newcommand{\markerglobaloptimum}{\raisebox{0.5pt}{\tikz{\node[draw,scale=0.4,cross=5pt,rotate=45,black](){};}}}
 \newcommand{\markerrobustoptimum}{\raisebox{0.5pt}{\tikz{\node[draw,scale=0.4,cross=5pt,black](){};}}}
 \newcommand{\markerdata}{\raisebox{0.5pt}{\tikz{\node[draw,scale=0.4,circle,black,fill=black](){};}}}
 \newcommand{\reddashed}{\raisebox{2pt}{\tikz{\draw[-,tableauC3,dotted,line width = 1.5pt](0,0) -- (5mm,0);}}}
 \newcommand{\redsolid}{\raisebox{2pt}{\tikz{\draw[-,tableauC3,line width = 1.5pt](0,0) -- (5mm,0);}}}
 \newcommand{\blackdashed}{\raisebox{2pt}{\tikz{\draw[-,black,dashed,line width = 1.5pt](0,0) -- (5mm,0);}}}
 \newcommand{\blacksolid}{\raisebox{2pt}{\tikz{\draw[-,black,line width = 1.5pt](0,0) -- (5mm,0);}}}
 \newcommand{\bluegp}{\raisebox{0pt}{\tikz{\draw[tableauC0!30!white,solid,fill=tableauC0!30!white,line width = 1.0pt](0.mm,0) rectangle (5.0mm,1.5mm);\draw[-,tableauC0,solid,line width = 1.0pt](0.,0.8mm) -- (5.0mm,0.8mm)}}}
 \newcommand{\greengp}{\raisebox{0pt}{\tikz{\draw[tableauC2!30!white,solid,fill=tableauC2!30!white,line width = 1.0pt](0.mm,0) rectangle (5.0mm,1.5mm);\draw[-,tableauC2,dashed,line width = 1.0pt](0.,0.8mm) -- (5.0mm,0.8mm)}}}
 \centering
 \begin{subfigure}[b]{.45\linewidth}
  \centering
  \fbox{\includegraphics[width=\linewidth]{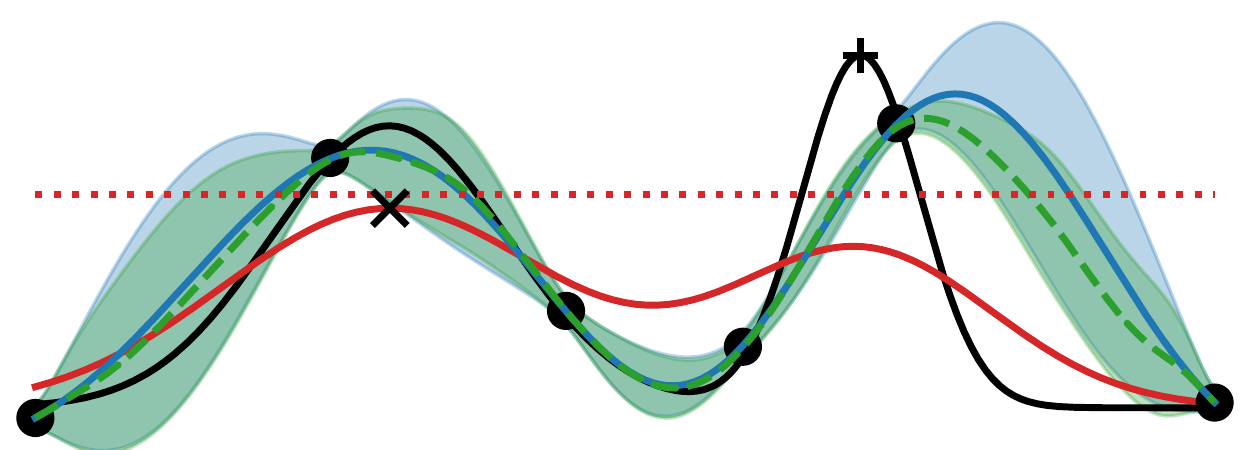}}
  \caption{Predictive distribution for $f(\bm{x})$ (\protect\blacksolid) before (\protect\bluegp) and after (\protect\greengp) conditioning on $g^*_k$ (\protect\reddashed).}
  \label{fig:fig_1_1d_schematic_gp_left}
 \end{subfigure}\hspace{2em}
 \begin{subfigure}[b]{.45\linewidth}
  \centering
  \fbox{\includegraphics[width=\linewidth]{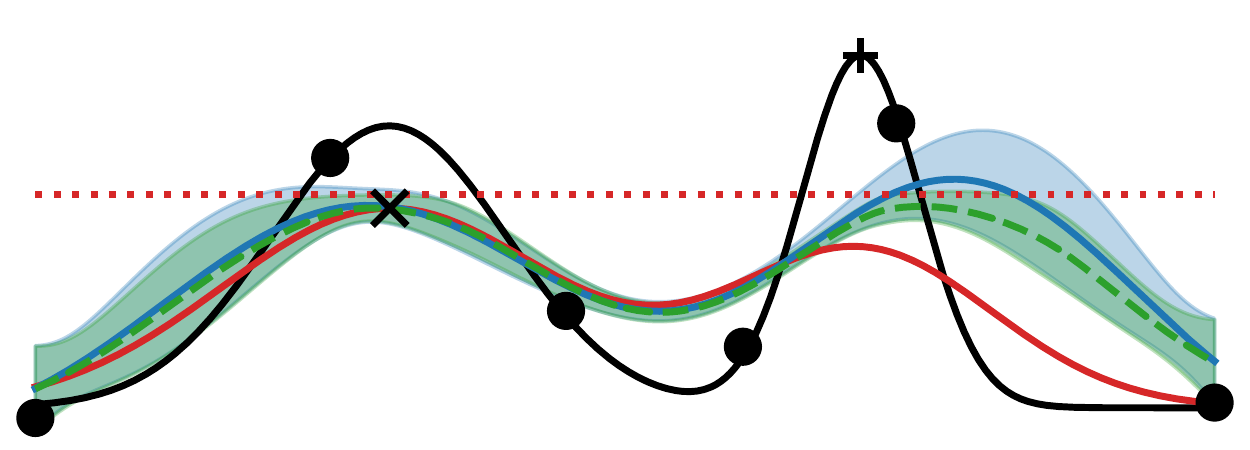}}
  \caption{Predictive distribution for $g(\bm{x})$ (\protect\redsolid) before (\protect\bluegp) and after (\protect\greengp) conditioning on $g^*_k$ (\protect\reddashed).}
  \label{fig:fig_1_1d_schematic_gp_right}
 \end{subfigure}
 \caption{Comparison of predictive distributions for the objective $f(\bm{x})$~(left) and the robust objective $g(\bm{x})$~(right) before and after conditioning on the sampled robust maximum value $g^*_k$. The goal is to find the robust maximum~(\protect\markerrobustoptimum) instead of the global maximum~(\protect\markerglobaloptimum); only $f(\bm{x})$ can be observed~(\protect\markerdata).}
 \label{fig:fig_1_1d_schematic_gp}
\end{figure*}
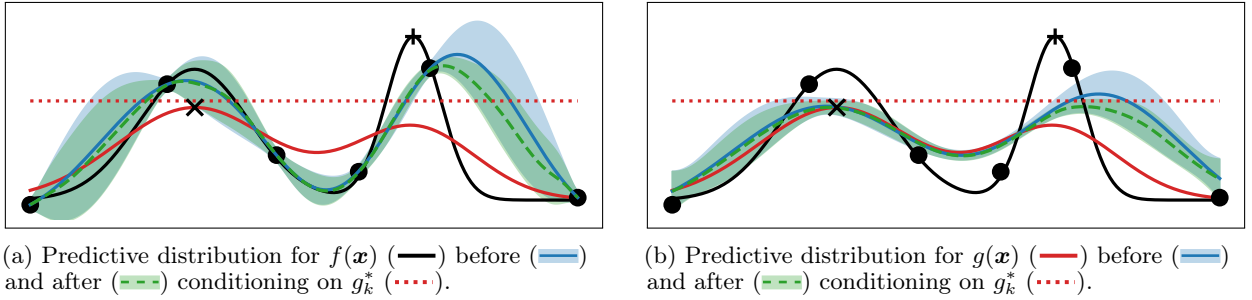

While no closed-form for $p(y(\bm{x}) | \datan, g^*_k)$ is known, it is straightforward to sample from this distribution via \acrfull{acr:rs}.
For the \gls{acr:rs}, a sampled function $\tilde{f}(\bm{x})$ from $p(y(\bm{x}) | \datan)$ is generated and its robust counterpart $\tilde{g}(\bm{x})$ is computed.
Given a robust maximum value sample $g^*_k \sim p(g^* | \datan)$, a sample is accepted when the maximum of $\tilde{g}(\bm{x})$ is smaller than $g^*_k$.
This process is repeated until $L$ samples have been accepted.
Pseudo-code for this procedure is shown in \lfalgref{alg:rejection_sampling_robust_max_value}.
Given the set $\tilde{Y}$ of $L$ accepted samples, we can approximate the entropy of the sample distribution as proposed by \citet{Ahmad1976NonparametricEntropyEstimation},
$H \big[ p( y(\bm{x}) | \datan, g^*_k ) \big] \approx - \frac{1}{L} \sum_{\tilde{y}_i \in \tilde{Y}} \ln \left[ \hat{p}(\tilde{y}_i) \right]$,
with $\hat{p}(\cdot)$ being the kernel density estimate of $p( y(\bm{x}) | \datan, g^*_k )$ based on $\tilde{Y}$ \citep{Rosenblatt1956KernelDensityEstimate}.
Note that $\hat{p}(\cdot)$ sums over all samples in $\tilde{Y}$, such that the entropy computation scales quadratically with $L$ due to a nested summation over $\tilde{Y}$.
In the experiments, we found that $L=1000$ samples result in a sufficiently accurate approximation.
Due to the Monte Carlo approximation in \lfeqref{eq:monte_carlo_approximation_expectation_max_vals}, the \gls{acr:rs} step is conducted for each $g^*_k \in G^*$, which renders the optimization of $\alpha_{\text{NES}}(\bm{x})$ costly.
In the following, we develop a more efficient approximation scheme.

\vspace{-5pt}
\subsubsection{Using Expectation Propagation}
\label{sec:ep_based_approximation}

For a computationally more efficient approximation of $H[p(y(\bm{x}) | \datan, g^*_k)]$, we exploit the fact that the entropy of a normal distribution is given analytically.
As the observation noise is additive, we can approximate the predictive distribution $p(f(\bm{x}) | \datan, g^*_k)$ and then add the observation noise to compute the entropy.
In the remainder of this section we discuss how to compute a Gaussian approximation to $p(f(\bm{x}) | \datan, g^*_k)$ with \gls{acr:ep}.
More details are given in the \version{appendix (\lfsecref{app:approximation_pred_distribution})}{supplementary material}.

We rewrite the conditioned posterior predictive distribution of $f(\bm{x})$ as
\begin{multline}\label{eq:predictive_distribution_rewritten}
p(f(\bm{x}) | \datan, g^*_k) = \\ \int p(f(\bm{x}) | \datan, g(\bm{x})) p(g(\bm{x}) | \datan, g^*_k) dg(\bm{x}).
\end{multline}
The first distribution, $p(f(\bm{x}) | \datan, g(\bm{x}))$, can be computed from \gls{acr:gp} arithmetic.
Note that the joint distribution $p(f(\bm{x}), \datan, g(\bm{x}))$ is a multivariate normal distribution and conditioning on $\datan$ and $g(\bm{x})$ results in $p(f(\bm{x}) | \datan, g(\bm{x})) = \mathcal{N}\left( f(\bm{x}) | \bm{\mu}_f, \bm{\Sigma}_f \right)$.
The second distribution in the integral, $p(g(\bm{x}) | \datan, g^*_k)$, is the predictive distribution for $g(\bm{x})$ with the constraint \mbox{$g(\bm{x}) \leq g^*_k$}.
This constraint can either be incorporated by a truncated normal distribution \citep{Wang2017MaxValueEntropySearch} or by a Gaussian approximation~\citep{Hoffman2015OPES}.
We follow the latter such that our approximation of the integral in \lfeqref{eq:predictive_distribution_rewritten} has an analytic solution:

\begin{enumerate}[leftmargin=0.5cm]
 \item The constraint \mbox{$g(\bm{x}) \leq g^*_k$} implies in particular that \mbox{$g(\bm{x}_i) \leq g^*_k$ for all $\bm{x}_i \in \datan$}, which results in a truncated normal distribution for $\bm{g} = [g(\bm{x}_1), \dots, g(\bm{x}_n)]^\top$.
 Aside from the univariate case, there exist no closed-form expressions for the mean and covariance of this distribution.
 We approximate the respective moments with \gls{acr:ep} \citep{Herbrich2005GaussianEP}, denoting the indicator function by $\indicator{\cdot}$,
 \begin{align}\label{eq:ep_approximation}
 p(\bm{g} | \datan, g^*_k) & \propto p(\bm{g} | \datan) \prod_{i=1}^{n} \indicator{\bm{x}_i | g(\bm{x}_i) \leq g^*_k} \notag \\
 & \stackrel{\text{(EP)}}{\approx} \mathcal{N} \left( \bm{g} | \bm{\mu}_1, \bm{\Sigma}_1 \right).  
 \end{align}
 
 \item By marginalizing out the latent function values $\bm{g}$, we obtain a predictive distribution.
 Deriving $p(g(\bm{x}) | \bm{g}, \datan)$ from \gls{acr:gp} arithmetic and substituting \lfeqref{eq:ep_approximation} results in
 \begin{align}\label{eq:unfinished_predictive_distribution}
 p_0(g(\bm{x}) | \datan, g^*_k) & = \int p(g(\bm{x}) | \bm{g}, \datan) p(\bm{g} | \datan, g^*_k) d\bm{g} \notag \\
 & \approx \mathcal{N}(g(\bm{x}) | m_0(\bm{x}), v_0(\bm{x})).  
 \end{align}
 \item Next, we incorporate the constraint that $g(\bm{x}) \leq g^*_k$ for all $\bm{x} \in \mathcal{X}$ by moment matching,
 \begin{align*}
 p(g(\bm{x}) | \datan, g^*_k) & \propto p_0(g(\bm{x}) | \datan, g^*_k) \indicator{\bm{x} | g(\bm{x}) \leq g^*_k} \\
 & \approx \mathcal{N}\left(g(\bm{x}) | \hat{m}(\bm{x}), \hat{v}(\bm{x})   \right).
 \end{align*}
 With the shorthand notation $\beta = (g^*_k - m_0(\bm{x})) / \sqrt{v_0(\bm{x})}$ and {$r = \varphi(\beta) / \Phi(\beta)$}, mean and variance are given by {$\hat{m}(\bm{x}) = m_0(\bm{x}) - \sqrt{v_0(\bm{x})} r$} and {$\hat{v}(\bm{x}) = v_0(\bm{x}) - v_0(\bm{x}) r (r + \beta)$} (see,~e.g.,~\citep{Jawitz2004TruncatedMoments}), where $\varphi(\cdot)$ and $\Phi(\cdot)$ denote the probability density function and cumulative density function of the standard normal distribution, respectively.
 The influence of conditioning $p(g(\bm{x}) | \datan)$ on the robust maximum value $g^*_k$ is visualized in \lffigref{fig:fig_1_1d_schematic_gp_right}.
 \item Approximating $p(g(\bm{x}) | \datan, g^*_k)$ with a Gaussian offers the benefit that the integral in \lfeqref{eq:predictive_distribution_rewritten} can be solved analytically as it is the marginalization over a product of Gaussians.
 Thus, the approximation to the posterior predictive distribution for $f(\bm{x})$ conditioned on $g^*_k$ is given by
 \begin{align*}
 p(f(\bm{x}) | \datan, g^*_k) \approx \mathcal{N}\left( f(\bm{x}) | \tilde{m}_k(\bm{x}), \tilde{v}_k(\bm{x}) \right).
 \end{align*}
 An exemplary visualization of this approximation is displayed in \lffigref{fig:fig_1_1d_schematic_gp_left}.
 Note that both, mean $\tilde{m}_k(\bm{x})$ and variance $\tilde{v}_k(\bm{x})$, are strongly influenced in regions of large predicted values.
\end{enumerate}
The final form of the \gls{acr:nes} acquisition function based on \gls{acr:ep} is then given by
\begin{multline}\label{eq:nes_ep}
\alpha_{\text{NES-EP}}(\bm{x}) =   \frac{1}{2} \bigg[ \log \Big( v_f(\bm{x} | \datan) + \sigma_\epsilon^2 \Big) \\ - \frac{1}{K} \sum_{g_k^* \in G^*} \log\Big( \tilde{v}_k(\bm{x}) + \sigma_\epsilon^2 \Big) \bigg].
\end{multline}
For each evaluation of \lfeqref{eq:nes_ep}, the variance $\tilde{v}_k(\bm{x})$ is computed for every sample $g^*_k$ separately.
The \gls{acr:ep} step iterates over all data points.
During experiments we found that it converges within 2--5 sweeps.
\lfeqref{eq:unfinished_predictive_distribution} dominates the computational cost due to the inversion of a kernel matrix of size $2n$, with $n$ being the number of data points.
The overall complexity of one evaluation is then $\mathcal{O}(Kn^3)$.
Please note that, unlike the \gls{acr:rs}-based approach that relies on a kernel density estimation, the entropy of the Gaussian approximation obtained with \gls{acr:ep} can be computed analytically.
In the following, we evaluate the proposed \gls{acr:nes} acquisition function and compare the \gls{acr:rs}- and \gls{acr:ep}-based approximations.

\section{Experiments}
In this section, we evaluate the \acrfull{acr:nes} acquisition function and compare it to other methods from the literature on a range of benchmark problems.
Furthermore, we consider an application from aerospace engineering, for which robustness of the design parameters is crucial.
For all experiments, we use a \gls{acr:se} kernel $k_f(\bm{x}, \bm{x}') = \sigma_f^2 \exp( -0.5 \norm{\bm{x} - \bm{x}'}^2_{\Lambda^{-1}} )$ with $\Lambda = \operatorname{diag}[\ell_1^2, \dots, \ell_{d}^2]$.
For this choice, $k_{gf}(\bm{x}, \bm{x}')$ and $k_g(\bm{x}, \bm{x}')$ (see \lfeqref{eq:predictive_distribution_g}) can be computed analytically.
As performance metric we choose the \gls{acr:ir} $r_n = | g(\bm{x}_n^*) - g^* |$, where $\bm{x}_n^*$ is the estimate of the robust optimum at iteration $n$.
For all experiments, we perform $100$ independent runs, each with different sets of initial points, albeit the same set across all methods.
The result figures show the median across all runs as line and the 25/75\textsuperscript{th} percentiles as shaded area.
The initial observations are uniformly sampled and the number of initial points is chosen depending on the dimensionality of the objective ($n_{0} = 3, 5, 10$ for $d = 1, 2, 3$, respectively).
We describe all evaluated methods below.
All approaches were implemented based on GPy \citep{Gpy2014} and the code to reproduce all results is publicly available at \url{https://github.com/boschresearch/NoisyInputEntropySearch}.

\newcommand{\axisheight}{0.9\linewidth}
\newcommand{\axiswidth}{1.05\linewidth}
\newcommand{\subfigurewidth}{.31\linewidth}
\newcommand{\captionvspace}{-2mm}

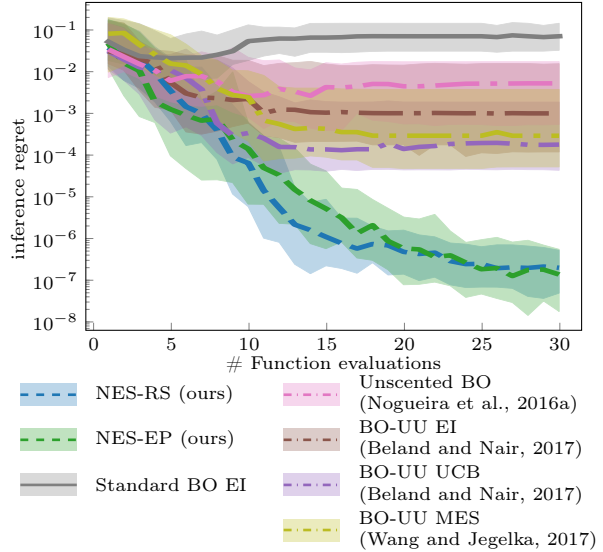
\begin{figure}
\centering
\begin{subfigure}[p]{.02\linewidth}
 {\figurefontsize\begin{tikzpicture}
\node[] at (0,0) {};
\node[rotate=90] at (0.0,1.0) {inference regret};
\end{tikzpicture}}
\end{subfigure}
\begin{subfigure}[p]{0.9\linewidth}
{\figurefontsize
\begin{tikzpicture}
\begin{axis}[
x label style={at={(0.5,-0.03)},anchor=south},
xmin=-0.5,xmax=31.4,
xlabel=\# Function evaluations,
ymode = log,
ylabel near ticks,
ymin=6.305e-9,ymax=4.575e-1,
major tick length=0.1cm,
minor tick length = 0.05cm,
tick pos=left,
height=0.8\linewidth,
width=1.1\linewidth,
axis on top,
max space between ticks=20
]
\addplot[thick,blue] graphics[xmin=-0.5,xmax=31.4,ymin=6.305e-9,ymax=4.575e-1,] {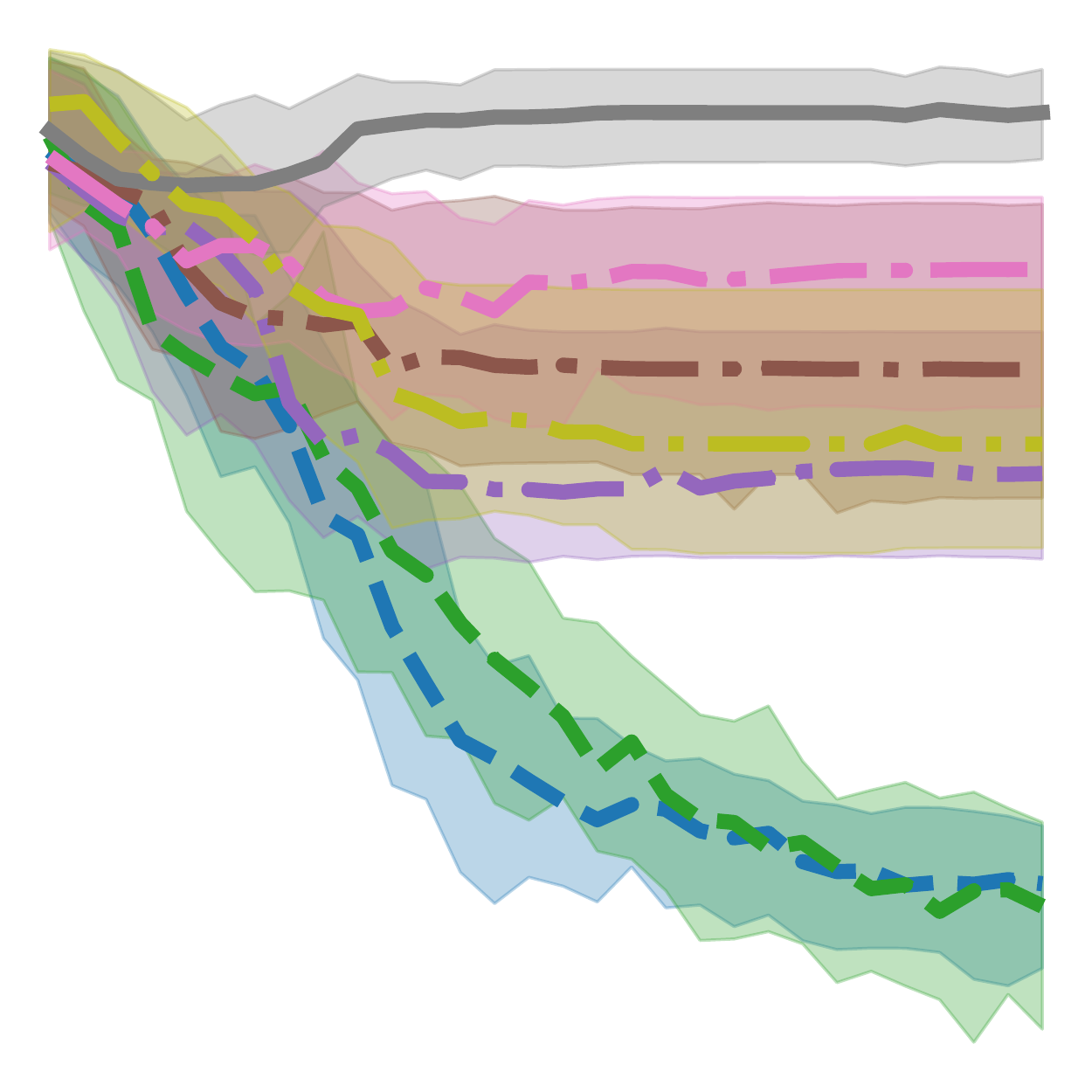};
\end{axis}
\end{tikzpicture}
}
\end{subfigure}
\begin{subfigure}[p]{\subfigurewidth}
 \vspace*{-0.6cm}
 \hspace*{-2.4cm}
 {\figurefontsize\begin{tikzpicture}

\def\barhalfheight{0.15}
\def\barwidth{0.75}
\def\dy{0.6}
\def\dxtext{0.7}
\def\dxbar{-0.2}
\def\dxcols{3.5}

\newcommand{\LegendListShort}{
 0/tableauC0/dashed/NES-RS (ours), 
 1/tableauC2/dashed/NES-EP (ours),
 2/tableauC7/solid/Standard BO EI}

\newcommand{\LegendListLong}{
 0/tableauC6/dashdotted/Unscented BO \\ \citep{Nogueira2016unscentedBO},
 1/tableauC5/dashdotted/BO-UU EI \\ \citep{Beland2017UncertaintyBONipsWorkshop},
 2/tableauC4/dashdotted/BO-UU UCB \\ \citep{Beland2017UncertaintyBONipsWorkshop},
 3/tableauC8/dashdotted/BO-UU MES \\ \citep{Wang2017MaxValueEntropySearch}}

\foreach \i/\markercolor/\linestyle/\entry in \LegendListShort {
 \node[anchor=west, align=left] at (\dxtext,-\i*\dy) {\entry};
 \fill [\markercolor!30!white] (\dxbar,\barhalfheight-\i*\dy) rectangle (\dxbar + \barwidth, -\barhalfheight-\i*\dy);
 \draw [-,\markercolor, \linestyle, line width = 1.0pt] (\dxbar, -\i*\dy) -- (\dxbar + \barwidth, -\i*\dy);
}

\foreach \i/\markercolor/\linestyle/\entry in \LegendListLong {
 \node[anchor=west, align=left] at (\dxtext + \dxcols,-\i*\dy) {\entry};
 \fill [\markercolor!30!white] (\dxbar + \dxcols,\barhalfheight-\i*\dy) rectangle (\dxbar + \barwidth + \dxcols, -\barhalfheight-\i*\dy);
 \draw [-,\markercolor, \linestyle, line width = 1.0pt] (\dxbar + \dxcols, -\i*\dy) -- (\dxbar + \barwidth + \dxcols, -\i*\dy);
}
\node at (0, 0) {};

\end{tikzpicture}}
\end{subfigure}
\vspace*{-0.5cm}
\caption{Within-model comparison in terms of the inference regret $r_n = |g(\bm{x}_n^*) - g^*|$. 
 We present the median (lines) and 25/75\textsuperscript{th} percentiles (shaded areas) across 50 different function samples from a \gls{acr:gp} prior.}
\label{fig:results_within_model_convergence}
\end{figure}

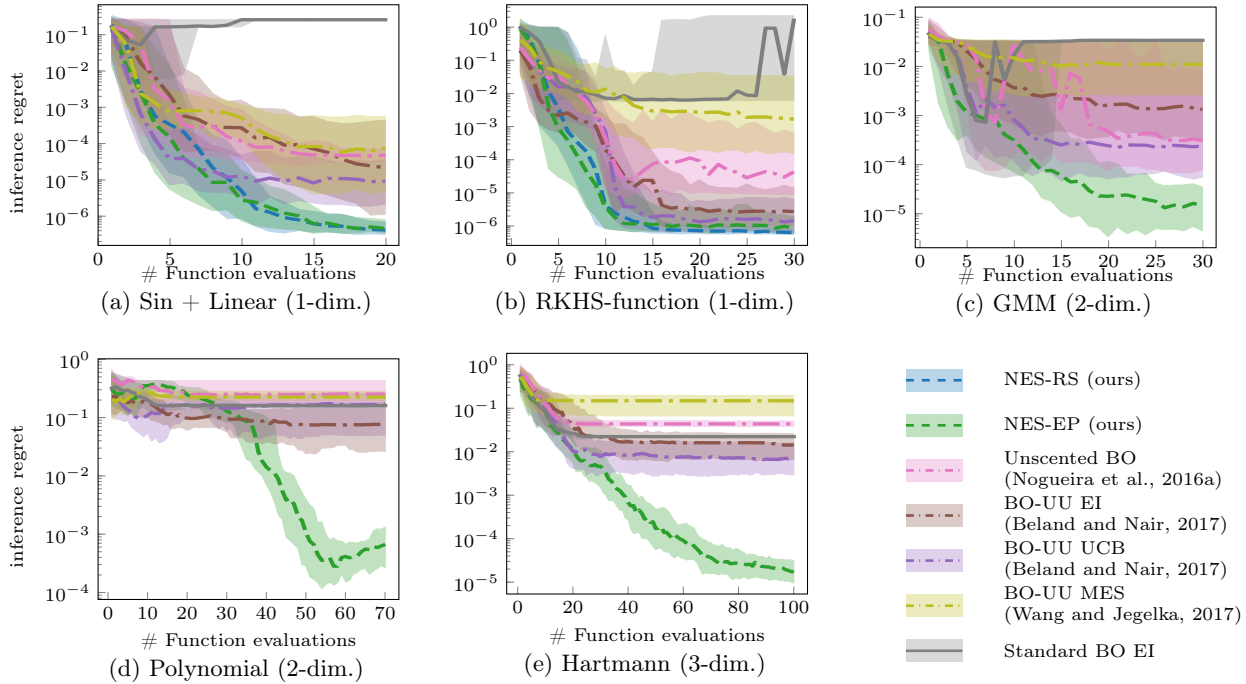
\begin{figure*}
\centering
\begin{subfigure}[p]{.02\linewidth}
 {\figurefontsize\begin{tikzpicture}
\node[] at (0,0) {};
\node[rotate=90] at (0.0,1.0) {inference regret};
\end{tikzpicture}}
\end{subfigure}
\begin{subfigure}[p]{\subfigurewidth}
{\figurefontsize
\begin{tikzpicture}
\begin{axis}[
x label style={at={(0.5,-0.02)},anchor=south},
xmin=0.0,xmax=20.9,
xlabel=\# Function evaluations,
ymode = log,
ylabel near ticks,
ymin=1.568e-7,ymax=6.682e-1,
major tick length=0.1cm,
minor tick length = 0.05cm,
tick pos=left,
height=\axisheight,
width=\axiswidth,
axis on top,
max space between ticks=20
]
\addplot[thick,blue] graphics[xmin=0.0,xmax=20.9,ymin=1.568e-7,ymax=6.682e-1,] {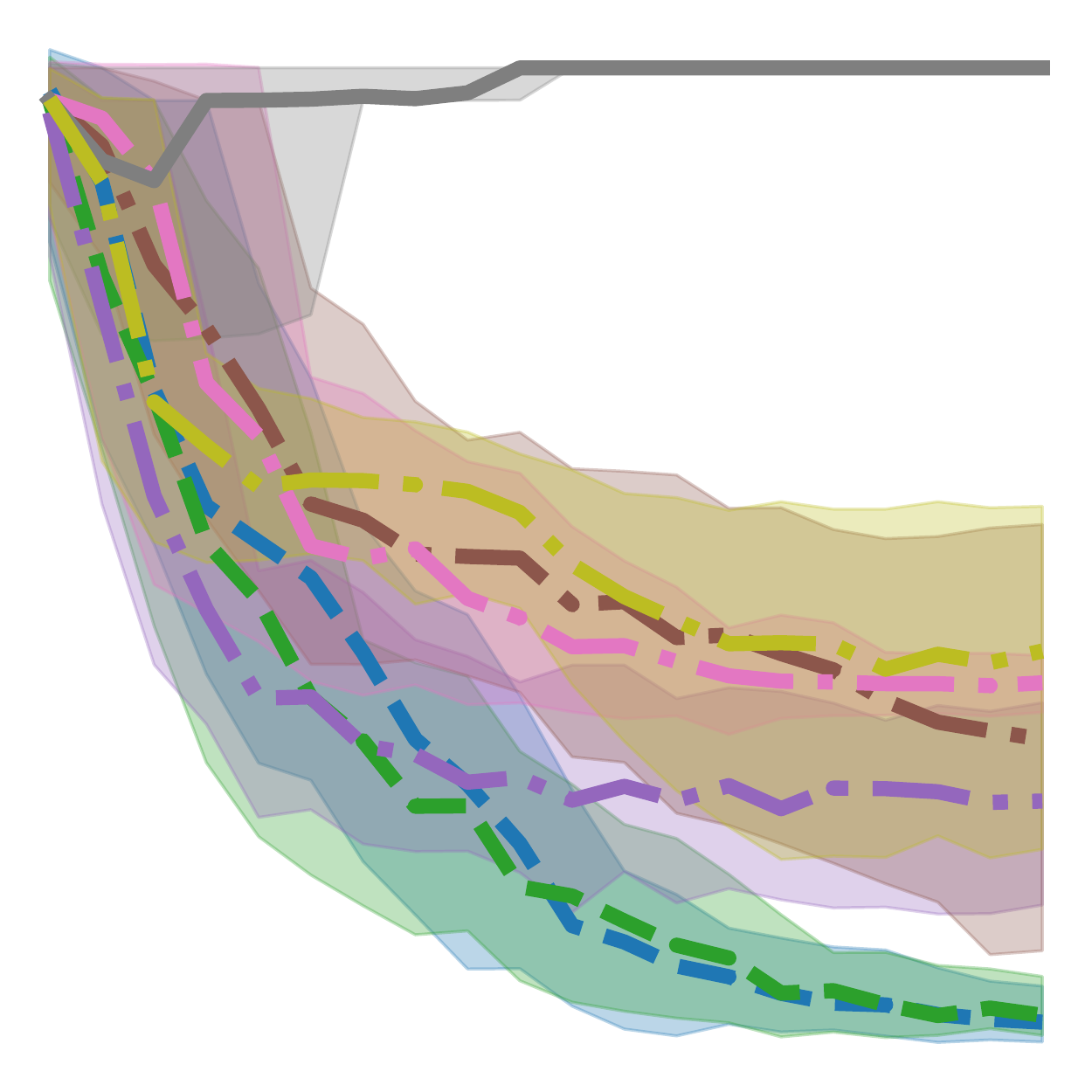};
\end{axis}
\end{tikzpicture}
}
\vspace*{\captionvspace}
 \caption{Sin + Linear (1-dim.)}\label{fig:results_convergence_synthetic_1d_01}
\end{subfigure}
\begin{subfigure}[p]{\subfigurewidth}
{\figurefontsize
 \begin{tikzpicture}
 \begin{axis}[
 x label style={at={(0.5,-0.02)},anchor=south},
 xmin=-0.5,xmax=31.4,
 xlabel=\# Function evaluations,
 ymode = log,
 ylabel near ticks,
 ymin=2.6e-7,ymax=4.749e-0,
 major tick length=0.1cm,
 minor tick length = 0.05cm,
 tick pos=left,
 height=\axisheight,
 width=\axiswidth,
 axis on top,
 max space between ticks=20
 ]
 \addplot[thick,blue] graphics[xmin=-0.5,xmax=31.4,ymin=2.6e-7,ymax=4.749e-0,] {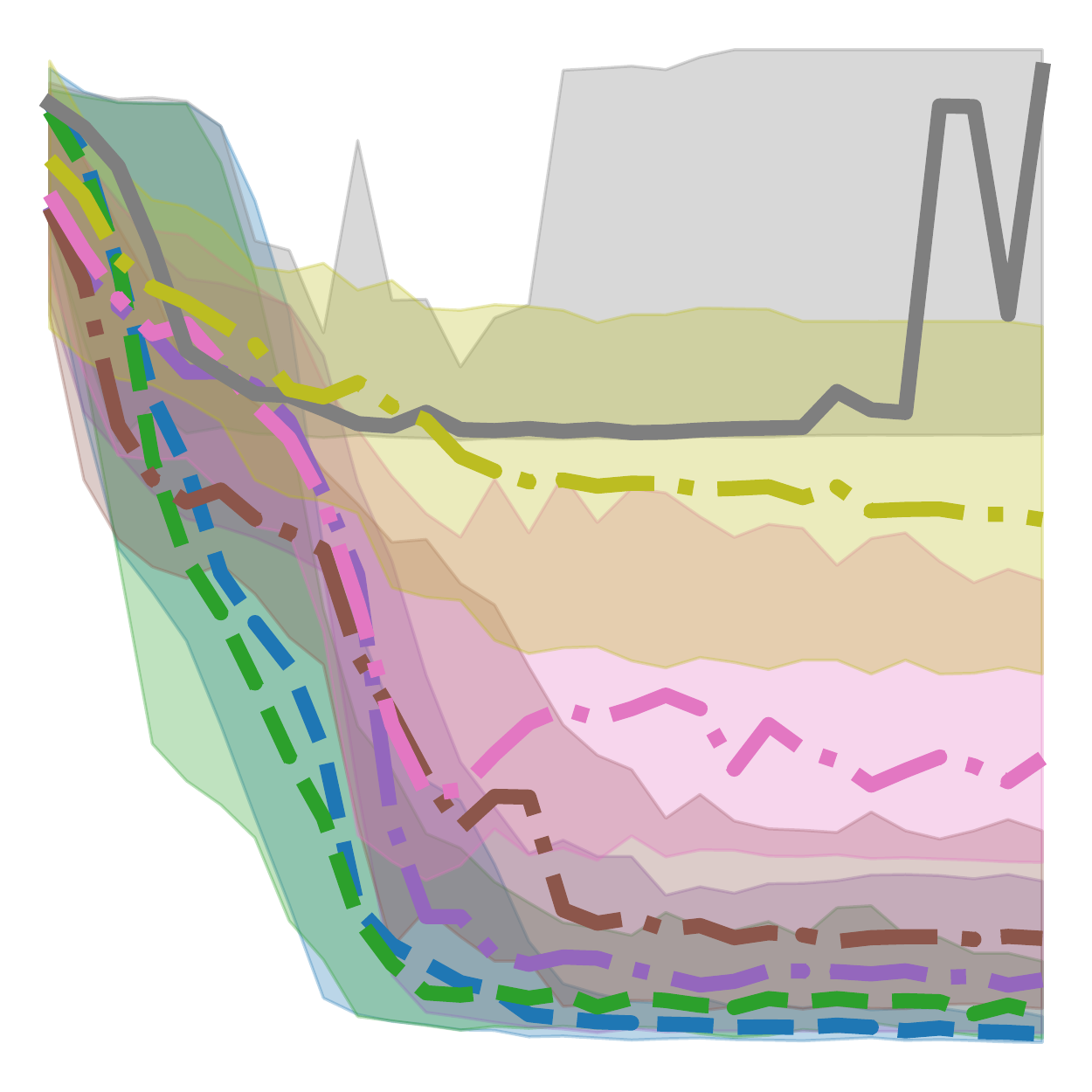};
 \end{axis}
 \end{tikzpicture}
}
\vspace*{\captionvspace}
 \caption{RKHS-function (1-dim.)}\label{fig:results_convergence_rkhs_synth}
\end{subfigure}
\begin{subfigure}[p]{\subfigurewidth}
{\figurefontsize
 \begin{tikzpicture}
 \begin{axis}[
 x label style={at={(0.5,-0.02)},anchor=south},
 xmin=-0.5,xmax=31.4,
 xlabel=\# Function evaluations,
 ymode = log,
 ylabel near ticks,
 ymin=2.034e-6,ymax=1.617e-1,
 major tick length=0.1cm,
 minor tick length = 0.05cm,
 tick pos=left,
 height=\axisheight,
 width=\axiswidth,
 axis on top,
 max space between ticks=20
 ]
 \addplot[thick,blue] graphics[xmin=-0.5,xmax=31.4,ymin=2.034e-6,ymax=1.617e-1,] {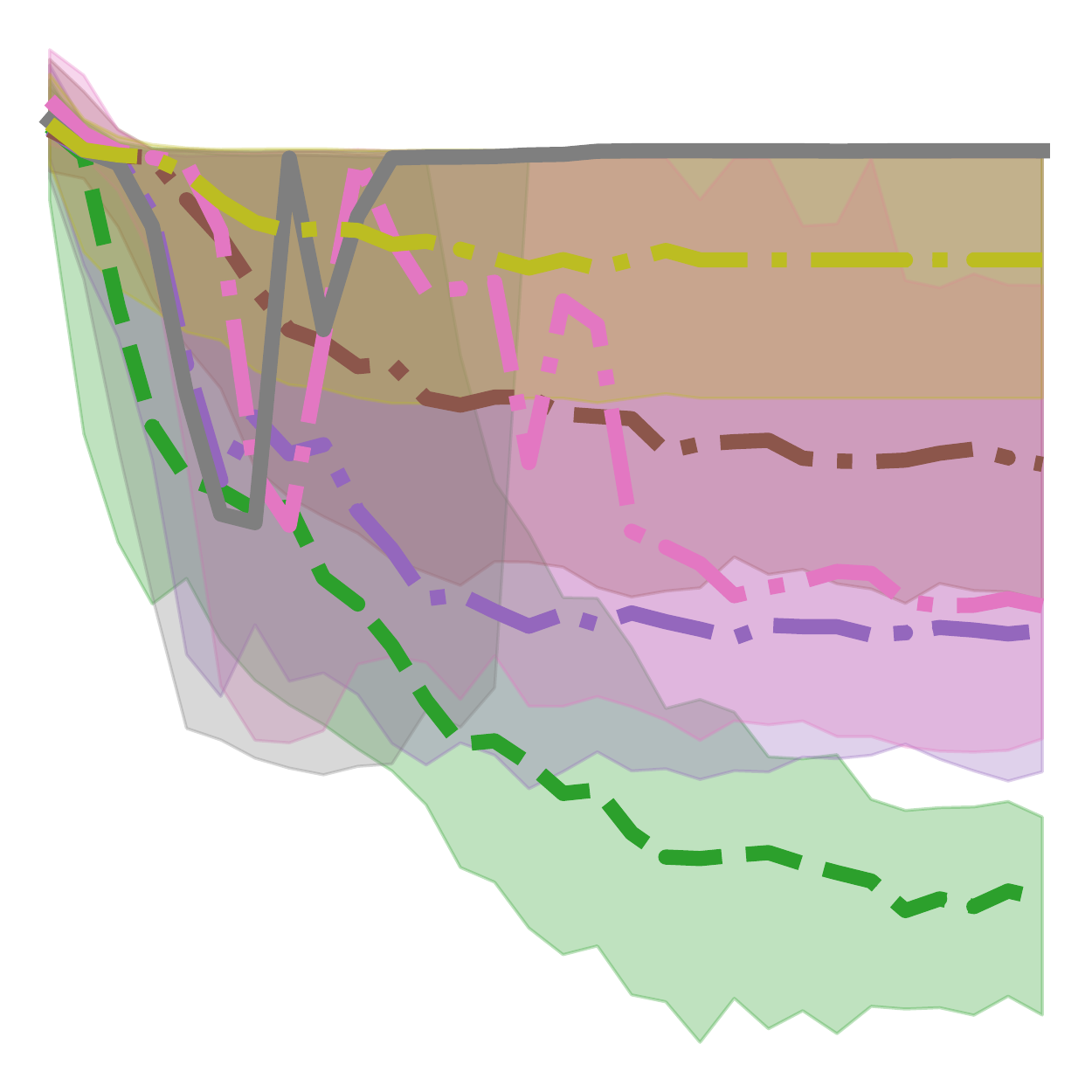};
 \end{axis}
 \end{tikzpicture}
}
\vspace*{\captionvspace}
 \caption{GMM (2-dim.)}\label{fig:results_convergence_gmm_2d}
\end{subfigure}

\vspace*{3mm}

\begin{subfigure}[p]{.02\linewidth}
 {\figurefontsize\begin{tikzpicture}
\node[] at (0,0) {};
\node[rotate=90] at (0.0,1.0) {inference regret};
\end{tikzpicture}}
\end{subfigure}
\begin{subfigure}[p]{\subfigurewidth}
 {\figurefontsize
  \begin{tikzpicture}
  \begin{axis}[
  x label style={at={(0.5,-0.02)},anchor=south},
  xlabel near ticks, 
  xmin=-2.5,xmax=73.5,
  xlabel=\# Function evaluations,
  ymode = log,
  ylabel near ticks,
  ymin=7.584e-5,ymax=1.019e0,
  major tick length=0.1cm,
  minor tick length = 0.05cm,
  tick pos=left,
  height=\axisheight,
  width=\axiswidth,
  axis on top,
  max space between ticks=15
  ]
  \addplot[thick,blue] graphics[xmin=-2.5,xmax=73.5,ymin=7.584e-5,ymax=1.019e0,] {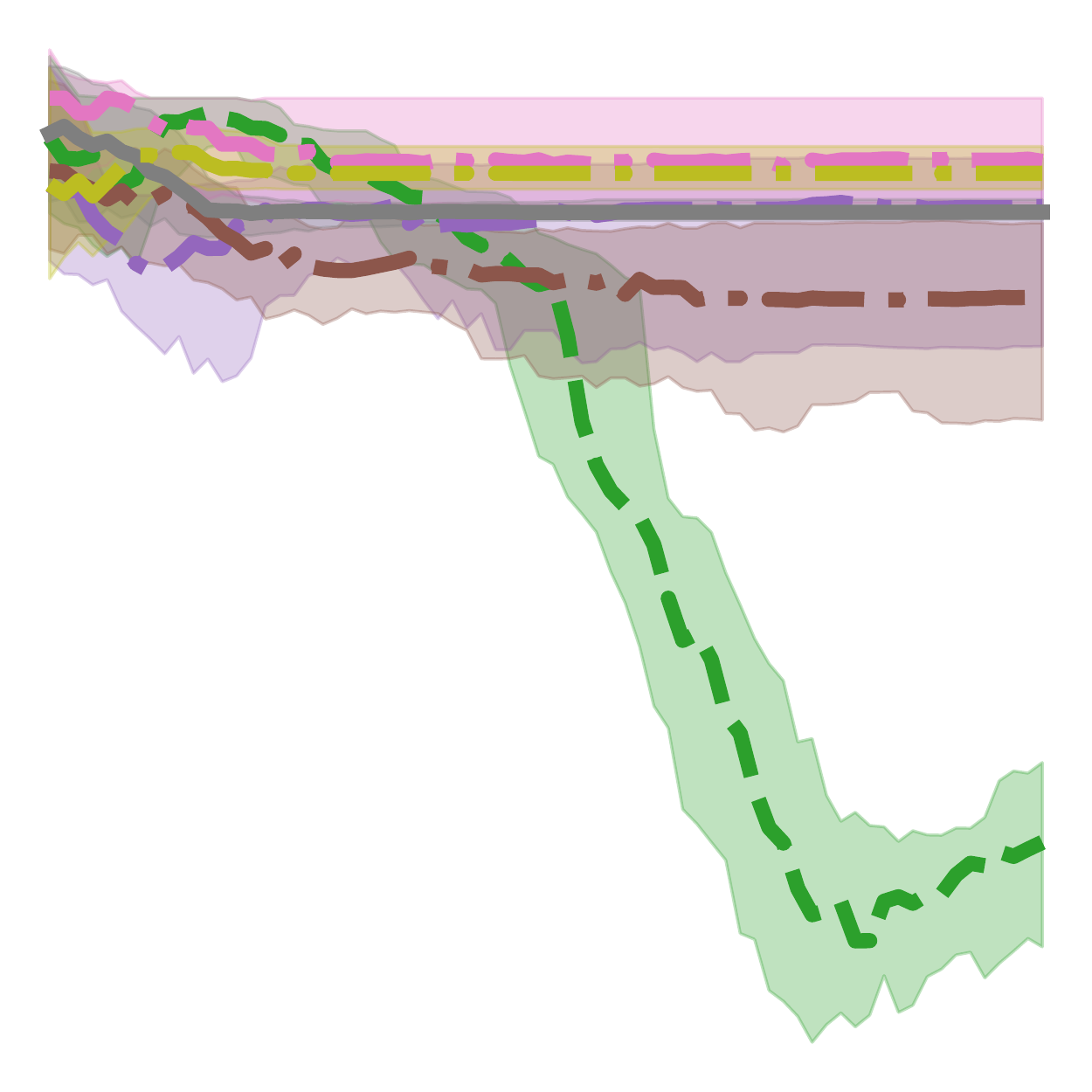};
  \end{axis}
  \end{tikzpicture}
 }
\vspace*{\captionvspace}
 \caption{Polynomial (2-dim.)}\label{fig:results_convergence_synth_poly_2d}
\end{subfigure}
\begin{subfigure}[p]{\subfigurewidth}
 {\figurefontsize
  \begin{tikzpicture}
  \begin{axis}[
  x label style={at={(0.5,-0.02)},anchor=south},
  xlabel near ticks, 
  xmin=-4.0,xmax=105.0,
  xlabel=\# Function evaluations,
  ymode = log,
  ylabel near ticks,
  ytick={1e-5, 1e-4, 1e-3, 1e-2, 1e-1, 1e0},
  ymin=5.341e-6,ymax=1.834e+0,
  major tick length=0.1cm,
  minor tick length = 0.05cm,
  tick pos=left,
  height=\axisheight,
  width=\axiswidth,
  axis on top,
  max space between ticks=20
  ]
  \addplot[thick,blue] graphics[xmin=-4.0,xmax=105.0,ymin=5.341e-6,ymax=1.834e+0,] {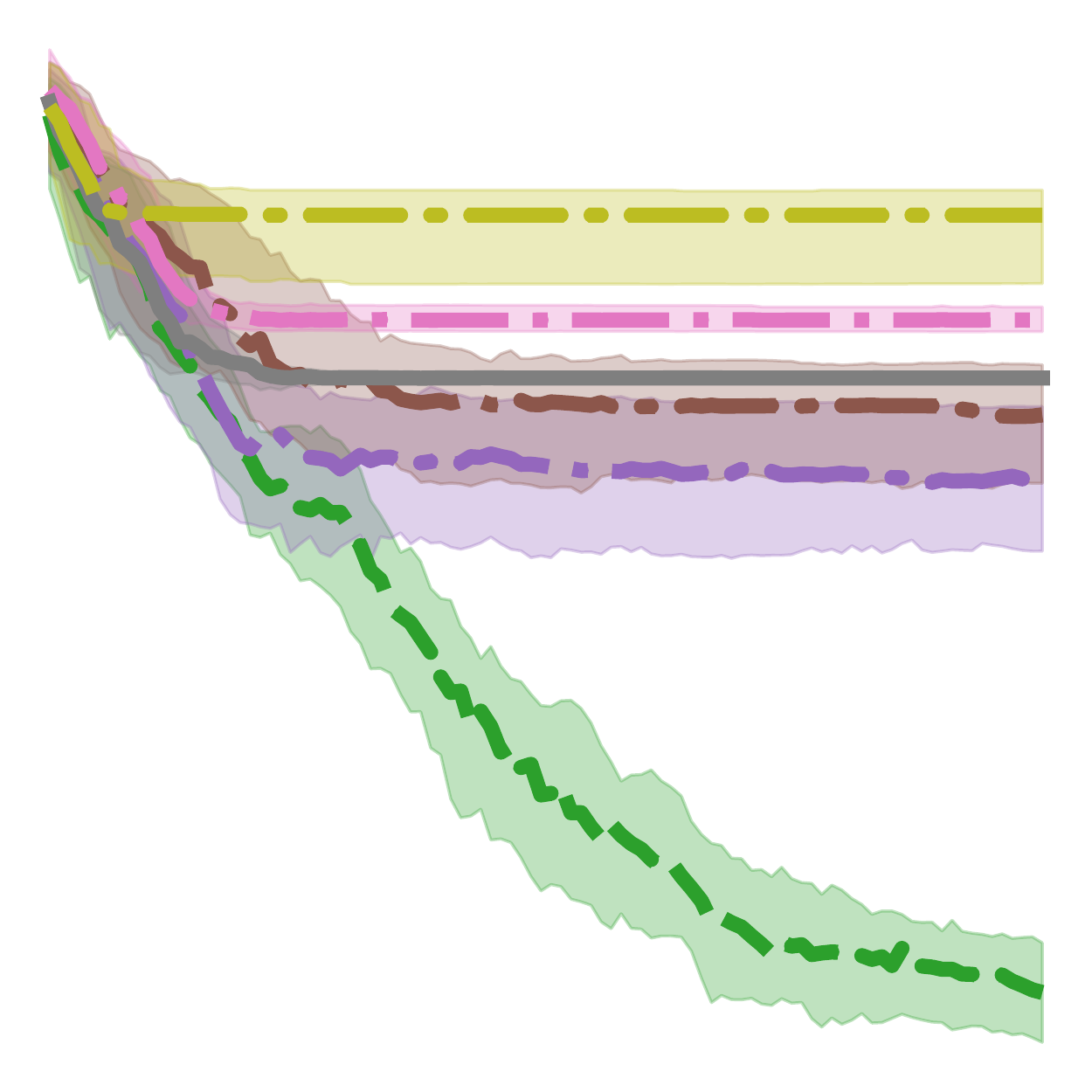};
  \end{axis}
  \end{tikzpicture}
 }
\vspace*{\captionvspace}
 \caption{Hartmann (3-dim.)}\label{fig:results_convergence_hartmann_3d}
\end{subfigure}
\begin{subfigure}[p]{\subfigurewidth}
 \vspace*{-0.0cm}
 \hspace*{+0.5cm}
 {\figurefontsize\begin{tikzpicture}

\def\barhalfheight{0.15}
\def\barwidth{0.75}
\def\dy{0.6}
\def\dxtext{1.2}
\def\dxbar{0.0}

\newcommand{\LegendList}{0/tableauC0/dashed/NES-RS (ours), 
                         1/tableauC2/dashed/NES-EP (ours),
                         2/tableauC6/dashdotted/Unscented BO \\ \citep{Nogueira2016unscentedBO},
                         3/tableauC5/dashdotted/BO-UU EI \\ \citep{Beland2017UncertaintyBONipsWorkshop},
                         4/tableauC4/dashdotted/BO-UU UCB \\ \citep{Beland2017UncertaintyBONipsWorkshop},
                         5/tableauC8/dashdotted/BO-UU MES \\
                         \citep{Wang2017MaxValueEntropySearch},
                         6/tableauC7/solid/Standard BO EI}

\node at (0.0, 0) {};
\foreach \i/\markercolor/\linestyle/\entry in \LegendList {
 \node[anchor=west, align=left] at (\dxtext,-\i*\dy) {\entry};
 \fill [\markercolor!30!white] (\dxbar,\barhalfheight-\i*\dy) rectangle (\dxbar + \barwidth, -\barhalfheight-\i*\dy);
 \draw [-,\markercolor, \linestyle, line width = 1.0pt] (\dxbar, -\i*\dy) -- (\dxbar + \barwidth, -\i*\dy);
}
\node at (0, -3.5) {};

\end{tikzpicture}}
\end{subfigure}
\caption{Inference regret $r_n = | g(\bm{x}_n^*) - g^* |$ on synthetic benchmark problems.
 We present the median (lines) and 25/75\textsuperscript{th} percentiles (shaded areas) across 100 independent runs with randomly sampled initial points. 
}
\label{fig:results_synthetic_functions_convergence}
\end{figure*}

\begin{itemize}[leftmargin=0.5cm]
 \itemsep-.2em 
 \item[-] \textbf{Noisy-Input Entropy Search (NES):} The proposed acquisition function using either rejection sampling (NES-RS)  or expectation propagation (NES-EP).
 For both variants of \gls{acr:nes}, we use \mbox{$M = 500$} random features for the \gls{acr:ssgp} and \mbox{$K = 1$} samples for the Monte Carlo estimate of the expectation over $p(g^* | \datan)$. 
 The number of accepted samples for \gls{acr:nes}-\gls{acr:rs} is set to \mbox{$L = 1000$}.
 \item[-] \textbf{BO Under Uncertainty (BO-UU):} The method presented by \citet{Beland2017UncertaintyBONipsWorkshop} which models the robust objective $g(\bm{x})$ as a \gls{acr:gp}, but assumes that it can be observed directly.
 We evaluate \mbox{BO-UU} with \acrfull{acr:ei}, \gls{acr:ucb} and \gls{acr:mes} \citep{Wang2017MaxValueEntropySearch}.
 \item[-] \textbf{Unscented BO:} The method presented by \citet{Nogueira2016unscentedBO} where the expectation over the input noise is approximated using an unscented transformation \citep{Julier2004UnscentedFiltering}.
 \item[-] \textbf{Standard \gls{acr:bo}:} Furthermore, we compare against standard \gls{acr:bo} with \gls{acr:ei} as acquisition function, which in general gives rise to non-robust optima.
\end{itemize}

\subsection{Within-Model Comparison}\label{sec:results_withon_model_comparison}
In a first step, we follow \citet{Hennig2012EntropySearch} and perform a within-model comparison.
For this analysis, we draw 50 function samples from a 1-dim. \gls{acr:gp} prior (SE-kernel with $\sigma_f = 0.5, \ell = 0.05$) and for each sample we try to find the robust optimum assuming the input noise $\sigma_x = 0.05$.
During optimization, the \gls{acr:gp} hyperparameters are fixed to their true values.
The benefit of this analysis is to isolate the influence of the acquisition functions from other factors such as the inference of hyperparameters or a potential model mismatch between objective and \gls{acr:gp} model.
For unknown objective functions, however, this procedure is not possible and the hyperparameters need to be inferred during optimization (see~\lfsecref{sec:results_synthetic_benchmark_functions}).
The results of the within-model comparison are presented in \lffigref{fig:results_within_model_convergence}.
Clearly, the two proposed acquisition functions \gls{acr:nes}-\gls{acr:rs} and \gls{acr:nes}-\gls{acr:ep} outperform all other approaches.
We observed that the \gls{acr:nes} acquisition functions continue to explore the vicinity of the robust optimum even at later stages of the optimization.
The other acquisition functions, however, stop exploring prematurely, which explains why the \gls{acr:ir}-curves level off early in \lffigref{fig:results_within_model_convergence}.
Furthermore, the performance of both \gls{acr:nes} variants is very similar in terms of \gls{acr:ir}, indicating that the \gls{acr:ep}-based approach is able to approximate the entropy terms in \lfeqref{eq:nes_acquisition_function} similarly well compared to the \gls{acr:rs}-based approach, but at lower computational cost.

\newcommand{\axisheightgravity}{5.0cm}
\newcommand{\axiswidthgravity}{4.0cm}
\newcommand{\subfigurewidthgravity}{0.25\linewidth}
\newcommand{\captionvspacegravity}{-2mm}

\begin{figure*}
 \begin{minipage}[t]{.28\linewidth}
  \centering
  
  {\figurefontsize
   \begin{tikzpicture}
   \begin{axis}[
   xlabel near ticks, 
   xmin=-10.0,xmax=45.0,
   xlabel=Initial angle $\alpha_0$,
   ylabel near ticks,
   ymin=4.1,ymax=4.99,
   ylabel=Initial speed $v_0$,
   major tick length=0.1cm,
   minor tick length = 0.05cm,
   tick pos=left,
   height=\axisheightgravity,
   width=\linewidth,
   axis on top,
   max space between ticks=20
   ]
   \addplot[thick,blue] graphics[xmin=-10.0,xmax=45.0,ymin=4.1,ymax=5.0,] {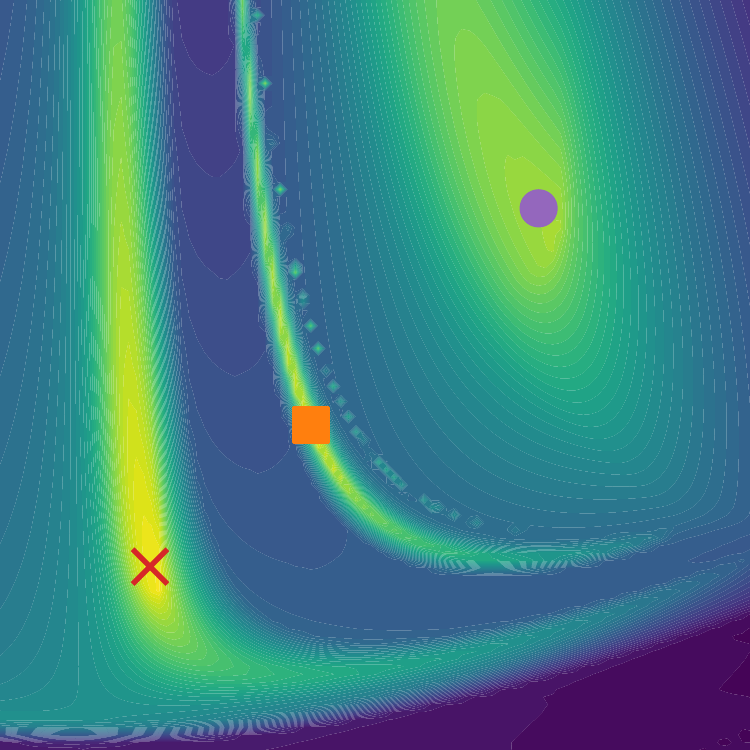};
   \end{axis}
   \end{tikzpicture}
  }
  
  \subcaption{Objective function}
  \label{fig:rocket_simulation_cost_function}
 \end{minipage}\hspace{-0.2cm}
 \begin{minipage}[b]{.05\linewidth}
  \centering
  {\figurefontsize
   \begin{tikzpicture}
   \begin{axis}[
   xmin=-10.0,xmax=45.0,
   xticklabels={,,},
   xmajorticks=false,
   ylabel near ticks,
   ytick={-1.2,-1.1,-1.0},
   ymin=-1.26,ymax=-0.93,
   major tick length=0.1cm,
   minor tick length = 0.05cm,
   tick pos=right,
   height=\axisheightgravity,
   width=2.5\linewidth,
   axis on top
   ]
   \addplot[thick,blue] graphics[xmin=-10.0,xmax=45.0,ymin=-1.26,ymax=-0.93,] {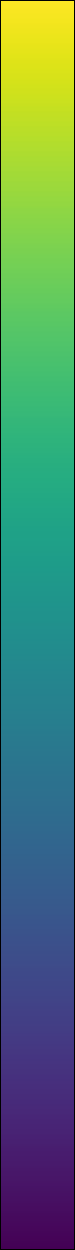};
   \end{axis}
   \node at (0.2, -0.28) {};
   \end{tikzpicture}
  }
  
 \end{minipage}%
 \begin{minipage}[t]{.33\linewidth}
  \centering
  {\figurefontsize
   \begin{tikzpicture}
   \begin{axis}[
   xlabel near ticks, 
   xmin=-0.5,xmax=10.5,
   xlabel=x-position,
   ylabel near ticks,
   ylabel shift=-3pt,
   ymin=-3.0,ymax=3.0,
   ylabel=y-position,
   major tick length=0.1cm,
   minor tick length = 0.05cm,
   tick pos=left,
   height=\axisheightgravity,
   width=\linewidth,
   axis on top,
   max space between ticks=20
   ]
   \addplot[thick,blue] graphics[xmin=-0.5,xmax=10.5,ymin=-3.0,ymax=3.0,] {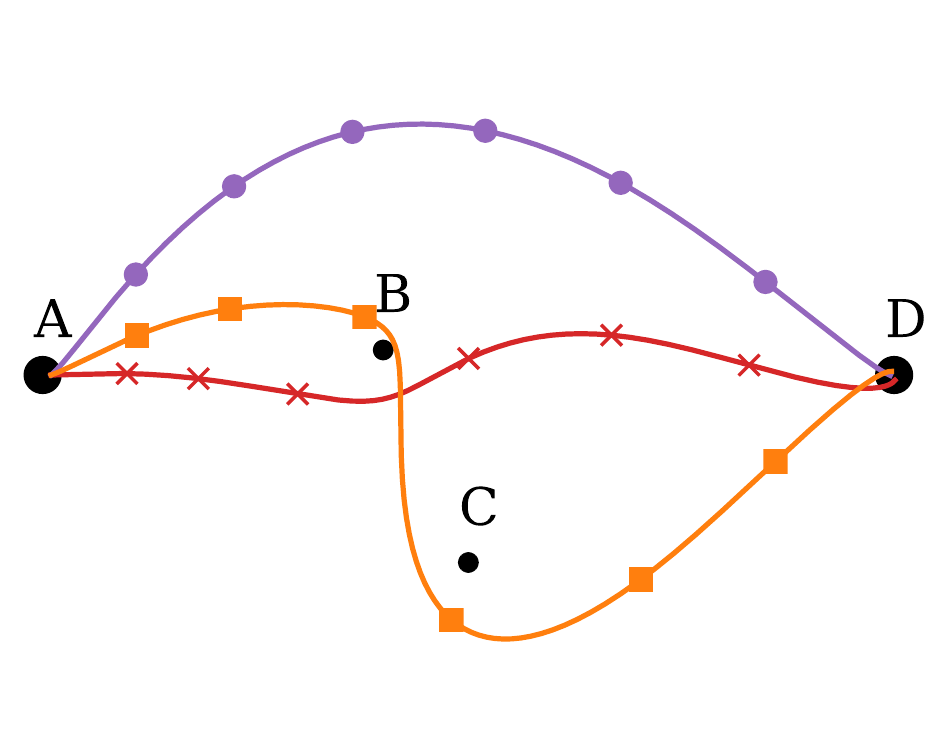};
   \end{axis}
   \end{tikzpicture}
  }
  
  \subcaption{Example trajectories}
  \label{fig:rocket_simulation_trajectories}
 \end{minipage}%
 \begin{minipage}[t]{.33\linewidth}
  \centering
  
  {\figurefontsize
   \begin{tikzpicture}
   \begin{axis}[
   xlabel near ticks, 
   xmin=-1.5,xmax=52.5,
   xlabel=\# Function evaluations,
   ymode = log,
   ylabel near ticks,
   ylabel shift=-3pt,
   ymin=2.3001e-4,ymax=2.0e-1,
   ylabel=inference regret,
   major tick length=0.1cm,
   minor tick length = 0.05cm,
   tick pos=left,
   height=\axisheightgravity,
   width=\linewidth,
   axis on top,
   max space between ticks=20,
   legend pos=north east,
   legend style={draw=none, fill=none}
   ]
   \addplot[dashed,line width=.5mm,tableauC2] graphics[xmin=-1.5,xmax=52.5,ymin=2.3001e-4,ymax=1.115e-1,] {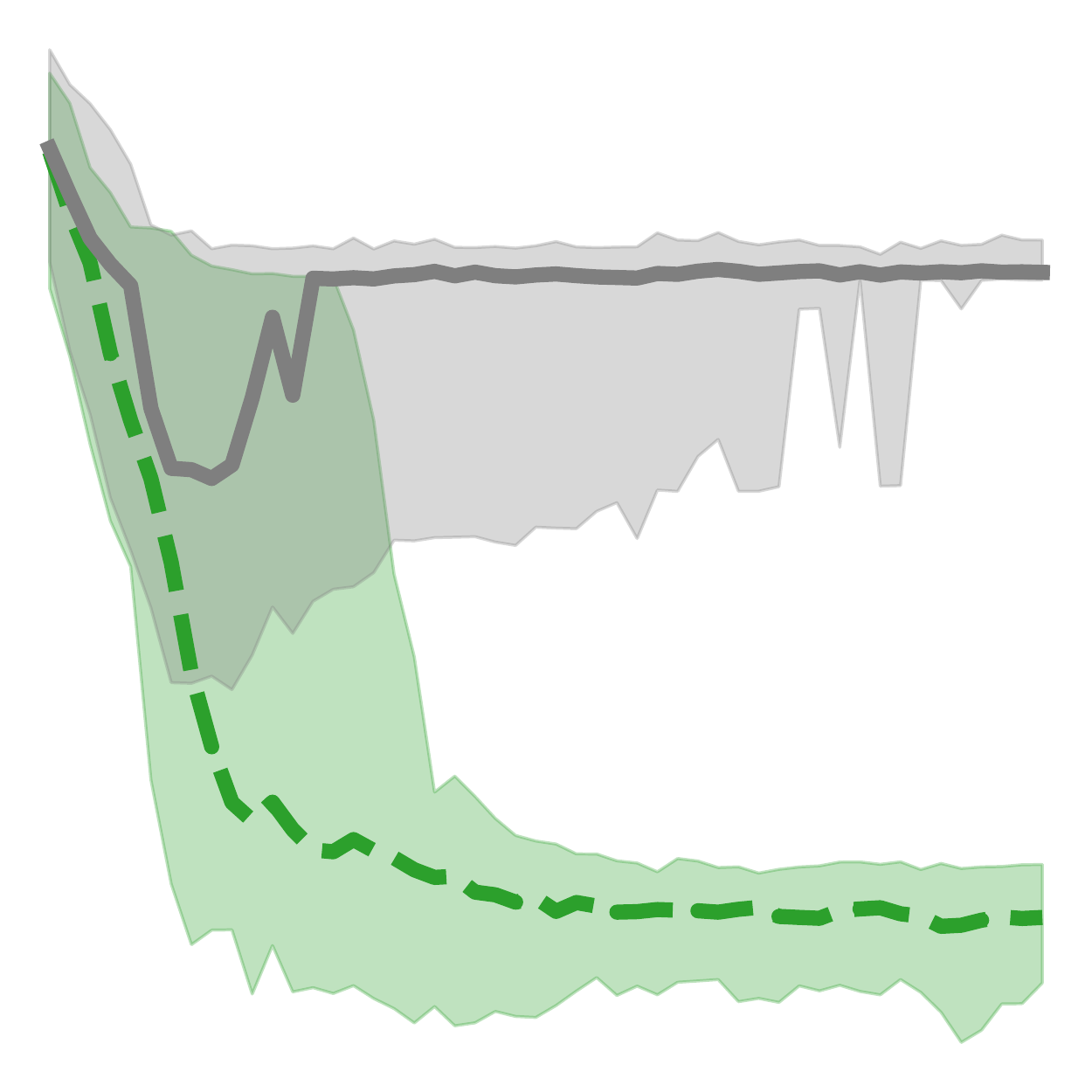};
   \addlegendentry{NES-EP (ours)};

   \addplot[line width=.5mm,tableauC7]{x};
   \addlegendentry{Standard BO EI};
   \end{axis}
   \end{tikzpicture}
  }
  
  \subcaption{Results}
  \label{fig:rocket_simulation_results}
 \end{minipage}
 \vspace*{\captionvspacegravity}
 \caption{Gravity assist maneuver: The goal of the maneuver is to get from planet A to planet D (see \lffigref{fig:rocket_simulation_trajectories}) by choosing an appropriate initial speed and starting angle. During the flight the engines are turned off such that all direction changes happen due to gravitational forces of planets A--D. The objective penalizes high values for the initial speed as well as the distance to the target planet, which results in the objective function as shown in \lffigref{fig:rocket_simulation_cost_function}. Results are shown in \lffigref{fig:rocket_simulation_results}. }
 \vspace*{\captionvspacegravity}
 \label{fig:rocket_simulation}
\end{figure*}

\vspace*{-2mm}
\subsection{Synthetic Benchmark Functions}\label{sec:results_synthetic_benchmark_functions}
\vspace*{-2mm}
We evaluate the aforementioned methods on the following functions:
\vspace*{-0.5em}
\begin{enumerate}[(a),leftmargin=0.7cm]
 \itemsep-.2em 
 \item $f(\bm{x}) = \sin(5 \pi \bm{x}^2) + 0.5 \bm{x}$, with $\bm{x} \in [0, 1]$ and input noise $\sigma_x = 0.05$,
 \item RKHS-function (1-dim.) \citep{Assael2014HeteroscedasticBO} with ${\sigma}_x = 0.03$, also used by \citet{Nogueira2016unscentedBO},
 \item Gaussian mixture model (2-dim.)  with $\bm{\Sigma}_x = 0.1^2 \bm{I}$, also used by \citet{Nogueira2016unscentedBO},
 \item Polynomial (2-dim.) \citep{Bertsimas2010RobustOptimization} with $\bm{\Sigma}_x = 0.6^2 \bm{I}$, also used by \citet{Bogunovic2018AdversariallyBO}.
 Here, we scaled and shifted the objective $f(x)$ s.t. $\mathbb{E}[f(x)] = 0.0$ and $\mathbb{V}[f(x)] = 1.0$,
 \item Hartmann (3-dim.)  with $\bm{\Sigma}_x = 0.1^2 \bm{I}$.
\end{enumerate}
Visualizations of the 1- and 2-dimensional functions are shown in \version{the appendix (\lffigref{fig:visualization_benchmarks})}{the supplementary material}.
The kernel hyperparameters as well as the observation noise are inferred via marginal likelihood maximization after each function evaluation.
Additionally, we chose a log-normal hyperprior for the kernel lengthscales, in order to relate them with the magnitude of the input noise which led to significantly more stable convergence for all acquisition functions.

In general, \gls{acr:nes} shows better convergence and \gls{acr:ir} across all benchmark functions, and this performance benefit increases with the dimensionality of the problem.
For all other methods that are designed to find the robust optimum, the performance is strongly task dependent.
Moreover, standard \gls{acr:bo} always finds the global optimum, which is, however, sub-optimal in the robust setting.
Note that the performance of standard \gls{acr:bo} appears to be competitive in terms of \gls{acr:ir} in \lffigref{fig:results_convergence_synth_poly_2d}/\ref{fig:results_convergence_hartmann_3d}, but the location of the estimated optimum is far off (see also \version{\lffigref{fig:dx_results_synthetic_functions_convergence} in the appendix}{the supplementary material} for an evaluation of the distance to the optimum).

\vspace*{-3mm}
\subsection{Application to Gravity Assist Maneuver}
\vspace*{-2mm}

We evaluate the proposed acquisition function on a so-called gravity assist maneuver.
The goal is to plan a spacecrafts trajectory to a target planet while minimizing energy consumption.
For this task gravitational effects from other planets are exploited in order to change the spacecraft's momentum without the need for active steering, thus saving fuel.
A visualization of this scenario is shown in \lffigref{fig:rocket_simulation_trajectories}.
The decision variables are 1)~$v_0$, the initial speed of the spacecraft and 2)~$\alpha_0$, the initial angle of flight from the start position.
The optimization objective for this task is given by $J(\alpha_0, v_0) = \log_{10} (d_{\text{target}} + \beta \cdot v_0)$, with $d_{\text{target}}$ being the closest distance of the resulting trajectory to the target planet and $\beta$ is a parameter that trades-off between the two cost terms.
The resulting cost function is depicted in \lffigref{fig:rocket_simulation_cost_function}, where the markers correspond to different resulting trajectories shown in \lffigref{fig:rocket_simulation_trajectories}.
The input noise is set to $\sigma_v = 0.05$ and $\sigma_\alpha = 3^\circ$ for $v_0$ and $\alpha_0$, respectively.
The results for \gls{acr:nes}-\gls{acr:ep} and standard \gls{acr:bo} (\gls{acr:ei}) are depicted in \lffigref{fig:rocket_simulation_results}.
Using either of the acquisition functions, the broad local optimum at the upper right corner of the domain (see~\lffigref{fig:rocket_simulation_cost_function}) is quickly explored within the first function evaluations.
After 10--15 function evaluations, standard \gls{acr:bo} finds the global optimum in the lower left corner and continues to exploit this region of the domain.
However, the global optimum is sensitive to perturbations on $\alpha_0$ and thus the inference regret stagnates.
On the other hand, \gls{acr:nes}-\gls{acr:ep} reliably finds the robust optimum and continues to explore the vicinity.
As a result, the inference regret is almost two orders of magnitude smaller compared to standard \gls{acr:bo}.

\vspace{-3mm}
\section{Conclusion}
\vspace{-3mm}
In this paper, we introduced a novel information-theoretic acquisition function for robust Bayesian optimization.
Our method, \acrfull{acr:nes}, considers a probabilistic formulation of the robust objective and maximizes the information gain about the robust maximum value~$g^*$.
Evaluation of \gls{acr:nes} requires the computation of the \gls{acr:gp}'s predictive distribution conditioned on the robust maximum value.
As this distribution is analytically intractable, we propose two approximation schemes.
The first is based on rejection sampling and is exact in the limit of infinite samples, but computationally challenging.
For the second approximation scheme we employ expectation propagation, which is computationally more efficient.
\gls{acr:nes} outperforms existing methods from the literature on a range of benchmark problems.
Finally, we demonstrated the practical importance of the proposed approach on a task from aerospace engineering 
where robustness is critical.

\subsubsection*{Acknowledgements}
The research of Melanie N. Zeilinger was supported by the Swiss National Science Foundation under grant no. PP00P2 157601/1.

\bibliographystyle{plainnat}
\bibliography{../../../Bibliography/conf_names_long,../../../Bibliography/library}

\begin{thebibliography}{42}
\providecommand{\natexlab}[1]{#1}
\providecommand{\url}[1]{\texttt{#1}}
\expandafter\ifx\csname urlstyle\endcsname\relax
  \providecommand{\doi}[1]{doi: #1}\else
  \providecommand{\doi}{doi: \begingroup \urlstyle{rm}\Url}\fi

\bibitem[Adida and Perakis(2006)]{Adida2006RobustOptimizationLogistics}
Elodie Adida and Georgia Perakis.
\newblock A robust optimization approach to dynamic pricing and inventory
  control with no backorders.
\newblock \emph{Mathematical Programming}, 107\penalty0 (1-2):\penalty0
  97--129, 2006.

\bibitem[Ahmad and Lin(1976)]{Ahmad1976NonparametricEntropyEstimation}
Ibrahim~A. Ahmad and Pi-Erh Lin.
\newblock A nonparametric estimation of the entropy for absolutely continuous
  distributions.
\newblock \emph{{IEEE} Transactions on Information Theory}, 22\penalty0
  (3):\penalty0 372--375, 1976.

\bibitem[Assael et~al.(2014)Assael, Wang, Shahriari, and
  de~Freitas]{Assael2014HeteroscedasticBO}
John-Alexander~M. Assael, Ziyu Wang, Bobak Shahriari, and Nando de~Freitas.
\newblock Heteroscedastic treed {B}ayesian optimisation.
\newblock \emph{arXiv preprint:1410.7172}, 2014.

\bibitem[Ba{\c{s}}ar and Bernhard(2008)]{Bacsar2008HInfinityControl}
Tamer Ba{\c{s}}ar and Pierre Bernhard.
\newblock \emph{H-infinity optimal control and related minimax design problems:
  a dynamic game approach}.
\newblock Springer Science \& Business Media, 2008.

\bibitem[Beland and Nair(2017)]{Beland2017UncertaintyBONipsWorkshop}
Justin~J. Beland and Prasanth~B. Nair.
\newblock Bayesian optimization under uncertainty.
\newblock \emph{{NIPS} Workshop on {B}ayesian Optimization}, 2017.

\bibitem[Bertsimas et~al.(2010)Bertsimas, Nohadani, and
  Teo]{Bertsimas2010RobustOptimization}
Dimitris Bertsimas, Omid Nohadani, and Kwong~Meng Teo.
\newblock Robust optimization for unconstrained simulation-based problems.
\newblock \emph{Operations Research}, 58\penalty0 (1):\penalty0 161--178, 2010.

\bibitem[Beyer and Sendhoff(2007)]{Beyer2007RobustOptimizationSurvey}
Hans-Georg Beyer and Bernhard Sendhoff.
\newblock Robust optimization--a comprehensive survey.
\newblock \emph{Computer Methods in Applied Mechanics and Engineering},
  196\penalty0 (33-34):\penalty0 3190--3218, 2007.

\bibitem[Bogunovic et~al.(2018)Bogunovic, Scarlett, Jegelka, and
  Cevher]{Bogunovic2018AdversariallyBO}
Ilija Bogunovic, Jonathan Scarlett, Stefanie Jegelka, and Volkan Cevher.
\newblock Adversarially robust optimization with {G}aussian processes.
\newblock In \emph{Advances in Neural Information Processing Systems ({NIPS})},
  pages 5765--5775, 2018.

\bibitem[Brochu et~al.(2010)Brochu, Cora, and de~Freitas]{Brochu2010TutorialBO}
Eric Brochu, Vlad~M. Cora, and Nando de~Freitas.
\newblock A tutorial on {B}ayesian optimization of expensive cost functions,
  with application to active user modeling and hierarchical reinforcement
  learning.
\newblock \emph{arXiv preprint:1012.2599}, 2010.

\bibitem[Calandra et~al.(2016)Calandra, Seyfarth, Peters, and
  Deisenroth]{Calandra2016GaitOptimization}
Roberto Calandra, Andr{\'e} Seyfarth, Jan Peters, and Marc~Peter Deisenroth.
\newblock Bayesian optimization for learning gaits under uncertainty.
\newblock \emph{Annals of Mathematics and Artificial Intelligence}, 76\penalty0
  (1-2):\penalty0 5--23, 2016.

\bibitem[Chen et~al.(2017)Chen, Lucier, Singer, and
  Syrgkanis]{Chen2017RobustNonConvexObjectives}
Robert~S. Chen, Brendan Lucier, Yaron Singer, and Vasilis Syrgkanis.
\newblock Robust optimization for non-convex objectives.
\newblock In \emph{Advances in Neural Information Processing Systems ({NIPS})},
  pages 4705--4714, 2017.

\bibitem[Chen et~al.(1996)Chen, Allen, Tsui, and
  Mistree]{Chen1996RobustDesignEngineering}
Wei Chen, Janet~K. Allen, Kwok-Leung Tsui, and Farrokh Mistree.
\newblock A procedure for robust design: minimizing variations caused by noise
  factors and control factors.
\newblock \emph{Journal of Mechanical Design}, 118\penalty0 (4):\penalty0
  478--485, 1996.

\bibitem[Cox and John(1992)]{Cox1992UpperConfidenceBound}
Dennis~D. Cox and Susan John.
\newblock A statistical method for global optimization.
\newblock In \emph{{IEEE} Transactions on Systems, Man, and Cybernetics}, pages
  1242--1246, 1992.

\bibitem[Cully et~al.(2015)Cully, Clune, Tarapore, and Mouret]{Cully2015nature}
Antoine Cully, Jeff Clune, Danesh Tarapore, and Jean-Baptiste Mouret.
\newblock Robots that can adapt like animals.
\newblock \emph{Nature}, 521\penalty0 (7553):\penalty0 503--507, 2015.

\bibitem[Dallaire et~al.(2009)Dallaire, Besse, and
  Chaib-Draa]{dallaire2009learning}
Patrick Dallaire, Camille Besse, and Brahim Chaib-Draa.
\newblock Learning {G}aussian process models from uncertain data.
\newblock In \emph{Proceedings of the International Conference on Neural
  Information Processing ({ICONIP})}, pages 433--440, 2009.

\bibitem[{GPy}(since 2012)]{Gpy2014}
{GPy}.
\newblock {GPy}: A {G}aussian process framework in python.
\newblock \url{http://github.com/SheffieldML/GPy}, since 2012.

\bibitem[Griffiths and
  Hern{\'a}ndez-Lobato(2017)]{Griffiths2017AutomaticChemicalDesignBO}
Ryan-Rhys Griffiths and Jos{\'e}~Miguel Hern{\'a}ndez-Lobato.
\newblock Constrained {B}ayesian optimization for automatic chemical design.
\newblock \emph{arXiv preprint:1709.05501}, 2017.

\bibitem[Groot et~al.(2010)Groot, Birlutiu, and
  Heskes]{Groot2010BayesianMonteCarloBO}
Perry Groot, Adriana Birlutiu, and Tom Heskes.
\newblock Bayesian {M}onte {C}arlo for the global optimization of expensive
  functions.
\newblock In \emph{Proceedings of the European Conference on Artificial
  Intelligence ({ECAI})}, pages 249--254, 2010.

\bibitem[Hennig and Schuler(2012)]{Hennig2012EntropySearch}
Philipp Hennig and Christian~J. Schuler.
\newblock Entropy search for information-efficient global optimization.
\newblock \emph{Journal of Machine Learning Research}, 13:\penalty0 1809--1837,
  2012.

\bibitem[Herbrich(2005)]{Herbrich2005GaussianEP}
Ralf Herbrich.
\newblock On {G}aussian expectation propagation.
\newblock Technical report, Microsoft Research Cambridge, 2005.

\bibitem[Hern{\'a}ndez-Lobato et~al.(2014)Hern{\'a}ndez-Lobato, Hoffman, and
  Ghahramani]{HernandezLobato2014PredictiveEntropySearch}
Jos{\'e}~Miguel Hern{\'a}ndez-Lobato, Matthew~W. Hoffman, and Zoubin
  Ghahramani.
\newblock Predictive entropy search for efficient global optimization of
  black-box functions.
\newblock In \emph{Advances in Neural Information Processing Systems ({NIPS})},
  pages 918--926, 2014.

\bibitem[Hoffman and Ghahramani(2015)]{Hoffman2015OPES}
Matthew~W. Hoffman and Zoubin Ghahramani.
\newblock Output-space predictive entropy search for flexible global
  optimization.
\newblock In \emph{{NIPS} Workshop on {B}ayesian Optimization}, 2015.

\bibitem[Jawitz(2004)]{Jawitz2004TruncatedMoments}
James~W. Jawitz.
\newblock Moments of truncated continuous univariate distributions.
\newblock \emph{Advances in Water Resources}, 27\penalty0 (3):\penalty0
  269--281, 2004.

\bibitem[Julier and Uhlmann(2004)]{Julier2004UnscentedFiltering}
Simon~J. Julier and Jeffrey~K. Uhlmann.
\newblock Unscented filtering and nonlinear estimation.
\newblock \emph{Proceedings of the {IEEE}}, 92\penalty0 (3):\penalty0 401--422,
  2004.

\bibitem[Kushner(1964)]{Kushner1964PI}
Harold~J. Kushner.
\newblock A new method of locating the maximum point of an arbitrary multipeak
  curve in the presence of noise.
\newblock \emph{Journal of Basic Engineering}, 86\penalty0 (1):\penalty0
  97--106, 1964.

\bibitem[L{\'a}zaro-Gredilla et~al.(2010)L{\'a}zaro-Gredilla,
  Qui{\~n}onero-Candela, Rasmussen, and
  Figueiras-Vidal]{LazaroGredilla2010SparseSpectrumGP}
Miguel L{\'a}zaro-Gredilla, Joaquin Qui{\~n}onero-Candela, Carl~Edward
  Rasmussen, and An{\'i}bal~R. Figueiras-Vidal.
\newblock Sparse spectrum {G}aussian process regression.
\newblock \emph{Journal of Machine Learning Research}, 11:\penalty0 1865--1881,
  2010.

\bibitem[Martinez-Cantin et~al.(2018)Martinez-Cantin, Tee, and
  McCourt]{Martinez2018OutlierBO}
Ruben Martinez-Cantin, Kevin Tee, and Michael McCourt.
\newblock Practical {B}ayesian optimization in the presence of outliers.
\newblock In \emph{Proceedings of the International Conference on Artificial
  Intelligence and Statistics ({AISTATS})}, pages 1722--1731, 2018.

\bibitem[Minka(2001)]{Minka2001ExpectationPropagation}
Thomas~P. Minka.
\newblock Expectation propagation for approximate {B}ayesian inference.
\newblock In \emph{Proceedings of the Conference on Uncertainty in Artificial
  Intelligence ({UAI})}, pages 362--369, 2001.

\bibitem[Mo{\v{c}}kus(1975)]{Mockus1975ExpectedImprovement}
Jonas Mo{\v{c}}kus.
\newblock On {B}ayesian methods for seeking the extremum.
\newblock In \emph{Optimization Techniques {IFIP} Technical Conference}, pages
  400--404, 1975.

\bibitem[Nogueira et~al.(2016{\natexlab{a}})Nogueira, Martinez-Cantin,
  Bernardino, and Jamone]{Nogueira2016unscentedBO}
Jos{\'e} Nogueira, Ruben Martinez-Cantin, Alexandre Bernardino, and Lorenzo
  Jamone.
\newblock Unscented {B}ayesian optimization for safe robot grasping.
\newblock In \emph{Proceedings of the {IEEE}/{RSJ} International Conference on
  Intelligent Robots and Systems ({IROS})}, pages 1967--1972,
  2016{\natexlab{a}}.

\bibitem[Nogueira et~al.(2016{\natexlab{b}})Nogueira, Martinez-Cantin,
  Bernardino, and Jamone]{Nogueira2016unscentedBOArxiv}
Jos{\'e} Nogueira, Ruben Martinez-Cantin, Alexandre Bernardino, and Lorenzo
  Jamone.
\newblock Unscented {B}ayesian optimization for safe robot grasping.
\newblock \emph{arXiv preprint arXiv:1603.02038}, 2016{\natexlab{b}}.

\bibitem[Oliveira et~al.(2019)Oliveira, Ott, and
  Ramos]{Oliveira2019UncertainInputBO}
Rafael Oliveira, Lionel Ott, and Fabio Ramos.
\newblock Bayesian optimisation under uncertain inputs.
\newblock In \emph{Proceedings of the International Conference on Artificial
  Intelligence and Statistics ({AISTATS})}, pages 1177--1184, 2019.

\bibitem[Rasmussen and Williams(2006)]{Rasmussen2006Book}
Carl~Edward Rasmussen and Christopher K.~I. Williams.
\newblock \emph{{G}aussian Processes for Machine Learning}.
\newblock MIT Press, 2006.

\bibitem[Rosenblatt(1956)]{Rosenblatt1956KernelDensityEstimate}
Murray Rosenblatt.
\newblock Remarks on some nonparametric estimates of a density function.
\newblock \emph{The Annals of Mathematical Statistics}, pages 832--837, 1956.

\bibitem[Sch{\"o}n and Lindsten(2011)]{Schoen2011ManipulatingMVGaussian}
Thomas~B. Sch{\"o}n and Fredrik Lindsten.
\newblock Manipulating the multivariate {G}aussian density.
\newblock Technical report, Link{\"o}ping University, 2011.

\bibitem[Shahriari et~al.(2016)Shahriari, Swersky, Wang, Adams, and
  de~Freitas]{Shahriari2016BayesianOptimization}
Bobak Shahriari, Kevin Swersky, Ziyu Wang, Ryan~P. Adams, and Nando de~Freitas.
\newblock Taking the human out of the loop: A review of {B}ayesian
  optimization.
\newblock \emph{Proceedings of the {IEEE}}, 104\penalty0 (1):\penalty0
  148--175, 2016.

\bibitem[Smith(2007)]{Smith2007MathematicsDFT}
Julius~Orion Smith.
\newblock \emph{Mathematics of the discrete {F}ourier transform ({DFT}): with
  audio applications}.
\newblock Julius Smith, 2007.

\bibitem[Snoek et~al.(2015)Snoek, Rippel, Swersky, Kiros, Satish, Sundaram,
  Patwary, Prabhat, and Adams]{Snoek2015DNGO}
Jasper Snoek, Oren Rippel, Kevin Swersky, Ryan Kiros, Nadathur Satish,
  Narayanan Sundaram, Mostofa Patwary, Mr~Prabhat, and Ryan Adams.
\newblock Scalable {B}ayesian optimization using deep neural networks.
\newblock In \emph{Proceedings of the International Conference on Machine
  Learning ({ICML})}, pages 2171--2180, 2015.

\bibitem[Srinivas et~al.(2010)Srinivas, Krause, Kakade, and
  Seeger]{Srinivas2010UpperConfidenceBound}
Niranjan Srinivas, Andreas Krause, Sham~M Kakade, and Matthias Seeger.
\newblock Gaussian process optimization in the bandit setting: No regret and
  experimental design.
\newblock In \emph{Proceedings of the International Conference on Machine
  Learning ({ICML})}, 2010.

\bibitem[Tesch et~al.(2011)Tesch, Schneider, and
  Choset]{Tesch2011SnakeGaitBOChangingEnvironments}
Matthew Tesch, Jeff Schneider, and Howie Choset.
\newblock Adapting control policies for expensive systems to changing
  environments.
\newblock In \emph{Proceedings of the {IEEE}/{RSJ} International Conference on
  Intelligent Robots and Systems ({IROS})}, pages 357--364. IEEE, 2011.

\bibitem[Toscano-Palmerin and Frazier(2018)]{Toscano2018ExpensiveIntegralBO}
Saul Toscano-Palmerin and Peter~I. Frazier.
\newblock Bayesian optimization with expensive integrands.
\newblock \emph{arXiv preprint:1803.08661}, 2018.

\bibitem[Wang and Jegelka(2017)]{Wang2017MaxValueEntropySearch}
Zi~Wang and Stefanie Jegelka.
\newblock Max-value entropy search for efficient {B}ayesian optimization.
\newblock In \emph{Proceedings of the International Conference on Machine
  Learning ({ICML})}, pages 3627--3635, 2017.

\end{thebibliography}

\version{\clearpage
 \appendix
 
\section{Expectation Over Input Noise for Sparse Spectrum GP Samples}\label{app:fourier_transform_ssgp}

Consider a sampled function from a \gls{acr:ssgp} of the form $\tilde{f}(\bm{x}) = \bm{a}^T \bm{\phi}_f (\bm{x})$.
In this section, we solve the following integral,
\begin{align}\label{eq:cross_correlation}
 \phi_{g, i}(\bm{x}) = \int \phi_{f, i}(\bm{x} + \bm{\xi}) p(\bm{\xi}) d\bm{\xi}
\end{align}
where $\phi_{f,i}(\bm{x})$ is the $i$-th component of $\bm{\phi}_{f}(\bm{x})$.
$\phi_{g, i}(\bm{x})$ is the $i$-th component of the corresponding 'robust' sample of the form $\tilde{g}(\bm{x}) = \bm{a}^T \bm{\phi}_g (\bm{x})$.
Note that the weights $\bm{a}$ are the same for both sampled functions, $\tilde{f}(\bm{x})$ and $\tilde{g}(\bm{x})$.

\lfeqref{eq:cross_correlation} requires the cross-correlation between function $\phi$ and $p$.
Since $p$ is a probability distribution (Gaussian in this case), it's complex conjugate is $p$ itself and the cross-correlation theorem states that in this case the cross-correlation is equivalent to the convolution \cite[Sec.~8.4]{Smith2007MathematicsDFT}.
Thus, we can apply the convolution theorem, which states
\begin{align*}
(\phi_{f,i} \ast p)(\bm{x}) = \mathcal{F}^{-1}\left\{ \mathcal{F} \left\{ \phi_{f,i} \right\} \mathcal{F} \left\{ p \right\} \right\},
\end{align*}
or in words: a convolution in 'time' domain is the same as a multiplication in frequency domain.
Before we apply this result, however, note that in the case of a separable filter window, we can apply the convolution in each dimension separately.
The final integral we need to solve then becomes,
\begin{align*}
\int \cos(\omega_{i,k} (x_k + \xi) + \underbrace{\sum_{j \neq k} \omega_{i, j} x_j + c_i}_{b_k}) p(\xi_k) d\xi_k,
\end{align*}
for $k = 1, \dots, n$.
We find the Fourier transforms of a shifted cosine with frequency $\omega_{i,k}$ and the univariate normal distribution, then multiply those and perform the inverse transform.
We use the following standard Fourier transforms:
\begin{multline*}
\mathcal{F}\left\{ \cos(\omega_{i, k} x_k + b_k) \right\} = \\
\sqrt{\frac{\pi}{2 }} \left( \delta(\omega - \omega_{i, k}) + \delta(\omega + \omega_{i, k})\right) \exp \left( j \frac{b_k}{\omega_{i, k}} \omega \right),
\end{multline*}
and
\begin{multline*}
\mathcal{F}\left\{ \frac{1}{\sqrt{2 \pi \sigma_{x,k}^2}} \exp \left( - \frac{x_k^2}{\sigma_{x,k}^2} \right) \right\}  = \\  \frac{1}{\sqrt{2 \pi}} \exp \left( - \frac{1}{2} \omega^2 \sigma_{x,k}^2 \right).
\end{multline*}
The inverse Fourier transform is given as
\begin{align*}
h(x) = \mathcal{F}^{-1}\left\{ \hat{h} \right\}(x) = \int \hat{h}(\omega) \exp(j\omega x) d\omega
\end{align*}
and plugging in the results from above gives
\begin{align*}
\begin{split}
\phi_{g,i}(\bm{x}) & = (\phi_{f,i} \ast p)(\bm{x}) \\
& = \phi_{f, i}(\bm{x})  \exp\Big( -\frac{1}{2} \sum_{j=1}^{d} \bm{w}_{i,j}^2 \sigma_{x,j}^2 \Big).
\end{split}
\end{align*}
Overall, filtering results in scaling of the basis functions.

\section{Details on EP-Approximation of the Conditional Predictive Distribution}\label{app:approximation_pred_distribution}

We aim at finding $p(f(\bm{x}) | \datan, g^*)$,
which is the predictive distribution for the latent function $f(\bm{x})$, i.e., the observable function, conditioned on the data $\datan$ and as well as on a sample of the robust maximum value distribution $g^*_k \sim p(g^* | \datan)$.
We will denote all evaluated points as $X = [\bm{x}_1, \dots, \bm{x}_n]$ and the corresponding observed function values as $\bm{y} = [y_1, \dots, y_n]$.

We start the derivation by rewriting the desired distribution as
\begin{multline}\label{eq:app_conditional_predictive_distribution}
p(f(\bm{x}) | \datan, g^*_k) = \int p(f(\bm{x}), g(\bm{x}) | \datan, g^*_k) dg(\bm{x}) \\
 = \int p(f(\bm{x}) | \datan, g(\bm{x})) p(g(\bm{x}) | \datan, g^*_k) dg(\bm{x}).
\end{multline}
We compute this integral in 3 steps:
First, we approximate $p(g(\bm{x}) | \datan, g^*_k)$ by a Gaussian distribution via \gls{acr:ep}.
Second, we compute $p(f(\bm{x}) | g(\bm{x}), \datan)$ by standard \gls{acr:gp} arithmetic.
Third, we make use of the fact that the marginalization over a product of Gaussian can be computed in closed form.

\paragraph{Gaussian approximation to $p(g(\bm{x}) | \datan, g^*_k)$:}
 We fit a Gaussian approximation to $p(g(\bm{x}) | \datan, g^*_k)$ as this enables us to compute the integral in \lfeqref{eq:app_conditional_predictive_distribution} in closed form.
 This approximation itself is done in three steps, following along the lines of \citet{Hoffman2015OPES} where they approximate $p(f(\bm{x}) | \datan, f^*)$.
 The key idea is that conditioning on the robust maximum value sample implies the constraint that $g(\bm{x}) \leq g^*$.
 \begin{enumerate}[leftmargin=0.5cm]
  \item In a first step, we only incorporate the constraint \mbox{$g(\bm{x}_i) \leq g_k^* \ \ \forall \ \ \bm{x}_i \in \datan $} such that
  \begin{align*}
  p(\bm{g} | \datan, g^*_k) \propto p(\bm{g} | \datan) \prod_{i=1}^{n}  \indicator{\bm{x}_i | g(\bm{x}_i) \leq g^*_k},
  \end{align*}
  where $\bm{g} = [g_1, \dots, g_n]$ denotes the latent function values of $g$ evaluated at $\bm{x}_1, \dots, \bm{x}_n$ and $\indicator{\cdot}$ is the indicator function.
  The above distribution constitutes a multi-variate truncated normal distribution.
  There is no analytical solution for its moments.
  One common strategy is to approximate the moments using \gls{acr:ep} \cite{Herbrich2005GaussianEP}.
  In practice, \gls{acr:ep} converges quickly for this distribution.
  We denote the outcome as
  \begin{align*}
  p(\bm{g} | \datan, g^*_k) \approx \mathcal{N}(\bm{g} | \bm{\mu}_1, \bm{\Sigma}_1).
  \end{align*}
  \item The next step is getting a predictive distribution from the (constrained) latent function values:
  \begin{align}\label{eq:app_marginal_affine_transformation_1}
  p_0(g(\bm{x}) | \datan, g^*_k) = \int p(\bm{g} | \datan, g^*_k) p(g(\bm{x}) | \datan, \bm{g})  d \bm{g}.
  \end{align}
  For the first term we use the Gaussian approximation of the previous step and the second term is given by standard GP arithmetic:
  \begin{align*}
  p(g(\bm{x}) | \datan, \bm{g}) & = \mathcal{N}(g(\bm{x}) | \bm{\mu}_g, \bm{\Sigma}_g),
  \end{align*}
  with
  \begin{align*}
  \begin{split}
  \bm{\mu}_g & = [ k_g(\bm{x}, X), k_{gf}(\bm{x}, X) ] \\ 
  & \qquad \begin{bmatrix}
  k_g(X, X) & k_{gf}(X, X) \\
  k_{fg}(X, X) & k_f(X, X) + \sigma_\epsilon^2 \bm{I}
  \end{bmatrix}^{-1}
  \begin{bmatrix}
  \bm{g} \\
  \bm{y}
  \end{bmatrix} \\
  & = [\bm{B}_1, \bm{B}_2]
  \begin{bmatrix}
  \bm{g} \\
  \bm{y}
  \end{bmatrix},
  \end{split}
  \end{align*}
  and
  \begin{align*}
  \bm{\Sigma}_g & = k_g(\bm{x}, \bm{x}) - [\bm{B}_1, \bm{B}_2]
  \begin{bmatrix}
  k_g(\bm{x}, X) \\
  k_{gf}(\bm{x}, X)
  \end{bmatrix}.
  \end{align*}
  Note that the integral in \lfeqref{eq:app_marginal_affine_transformation_1} is the marginalization over a product Gaussians where the mean of $p(g(\bm{x} | \datan, \bm{g})$ is an affine transformation of $\bm{g}$.
  Integrals of this form occur often when dealing with Gaussian distributions, e.g., in the context of Kalman filtering, and can be solved analytically (see e.g., \citet[Corollary 1]{Schoen2011ManipulatingMVGaussian}). 
  We obtain
  \begin{align*}
  p_0(g(\bm{x}) | \datan, g^*_k) \approx \mathcal{N}(g(\bm{x}) | m_0, v_0)),
  \end{align*}
  with 
  \begin{align*}
  m_0(\bm{x}) & = \bm{B}_1 \bm{\mu}_1 + \bm{B}_2 \bm{y}\\
  v_0(\bm{x}) & = \bm{\Sigma}_g + \bm{B}_1 \bm{\Sigma}_1 \bm{B}_1^T.
  \end{align*}
  \item Recall that in the first step we only enforced the constraints on the function values at the data points.
  Thus, we still need to integrate the constraint $g(\bm{x}) \leq g^*_k \ \ \forall \ \ \bm{x} \in \mathcal{X}$
  \begin{align*}
  p(g(\bm{x}) | \datan, g^*_k) \propto \mathcal{N}(m_0, v_0) \indicator{\bm{x} | g(\bm{x}) \leq g^*_k},
  \end{align*}
  where we again utilize a Gaussian approximation to this distribution.
  However, this is only a univariate truncated normal distribution and we can easily find the corresponding moments, such that
  \begin{align}\label{eq:app_tmp_1}
  p(g(\bm{x}) | \datan, g^*_k) \approx \mathcal{N}(g(\bm{x}) | \hat{m}(\bm{x}), \hat{v}(\bm{x})),
  \end{align}
  with mean and variance given as
  \begin{align*}
  \hat{m}(\bm{x}) & = m_0(\bm{x}) - \sqrt{v_0(\bm{x}) r}, \\
  \hat{v}(\bm{x}) & = v_0(\bm{x}) - v_0(\bm{x}) r ( r + \alpha),
  \end{align*}
  where $\alpha = (g^*_k - m_0(\bm{x})) / \sqrt{v_0(\bm{x})}$ and $r = \varphi(\alpha) / \Phi(\alpha)$. 
  As usual, $\varphi(\cdot)$ and $\Phi(\cdot)$ denote the PDF and CDF of the standard normal distribution, respectively.
 \end{enumerate}

\paragraph{\gls{acr:gp} arithmetic to find $p(f(\bm{x}) | g(\bm{x}), \bm{y})$:}

Starting with the joint distribution of all involved variables
\begin{align*}
\begin{split}
&\begin{bmatrix} f(\bm{x}) \\ \bm{y} \\ g(\bm{x}) \end{bmatrix} \sim  \mathcal{N}
\left(
\bm{0}, \bm{K}
\right),
\end{split} \\
& \bm{K} = \begin{bmatrix}
k_f(\bm{x}, \bm{x}) & k_f(\bm{x}, X)  & k_{fg}(\bm{x}, \bm{x})\\
k_f(X, \bm{x}) & k_f(X, X) + \sigma_n^2 I & k_{fg}(X, \bm{x}) \\
k_{gf}(\bm{x}, \bm{x}) & k_{gf}(\bm{x}, X) & k_g(\bm{x}, \bm{x})
\end{bmatrix},
\end{align*}

we introduce $\bm{z} = [\bm{y}, g(\bm{x})]^T$ for notational convenience and rewrite the joint distribution as
\begin{align}
\begin{bmatrix} f(\bm{x}) \\ \bm{z} \end{bmatrix} \sim \mathcal{N}
\left(
\bm{0},
\begin{bmatrix}
k_f(\bm{x}, \bm{x}) & k_z(\bm{x}, X)^T\\
k_z(\bm{x}, X) & K_z(\bm{x}, X)
\end{bmatrix}
\right).
\end{align}
Conditioning then gives
\begin{align}
p(f(\bm{x}) |& \bm{z})  = \mathcal{N} \left( f(\bm{x}) | \bm{\mu}_4 , \bm{\Sigma}_4 \right) \label{eq:app_tmp_2}\\
\bm{\mu}_4 & = k_z(\bm{x}, X)^T K_z(\bm{x}, X)^{-1} \bm{z} \notag \\
\bm{\Sigma}_4 & = k_f(\bm{x}, \bm{x}) - k_z(\bm{x}, X)^T K_z(\bm{x}, X)^{-1} k_z(\bm{x}, X) \notag.
\end{align}
Let's rewrite the mean of \lfeqref{eq:app_tmp_2} as follows
\begin{align}
\bm{\mu}_4 = \underbrace{k_z(\bm{x}, X)^T K_z(\bm{x}, X)^{-1}}_{ = [\bm{A}_1, \bm{A}_2]} \bm{z} = \bm{A}_1 \bm{y} + \bm{A}_2 g(\bm{x}),
\end{align}
with $\bm{A}_1$ and $\bm{A}_2$ being of appropriate dimensions.

\paragraph{Solve the integral:}
Now that we have the explicit forms of the distributions in the integral, we make use of the results \eqref{eq:app_tmp_1} and \eqref{eq:app_tmp_2},
 \begin{align}\label{eq:predictive_distribution_integral}
 & p(f(\bm{x}) | \datan, g^*_k) \\ & = \int p(f(\bm{x}) | \datan, g(\bm{x})) p(g(\bm{x}) | \datan, g^*_k) dg(\bm{x}) \\
 & = \int \mathcal{N}\left( f(\bm{x}) | \bm{A}_1 \bm{y} + \bm{A}_2 g(\bm{x}), \bm{\Sigma}_4 \right) \notag \\ 
 & \qquad \qquad \mathcal{N} \left( g(\bm{x}) | \hat{m}(\bm{x}), \hat{v}(\bm{x}) \right) dg(\bm{x}).
 \end{align}
 This integral has the same form as \lfeqref{eq:app_marginal_affine_transformation_1} and can be solved in closed form as well (see \cite[Corollary 1]{Schoen2011ManipulatingMVGaussian}).
 The final result is
 \begin{align}
 p(f(\bm{x}) | \datan, g^*_k) & \approx \mathcal{N}\left( f(\bm{x}) | \tilde{m}(\bm{x}), \tilde{v}(\bm{x}) \right) \\
 \tilde{m}(\bm{x}) & = \bm{A}_1 \bm{y} + \bm{A}_2 \hat{m}(\bm{x}) \\
 \tilde{v}(\bm{x}) & = \bm{\Sigma}_4 + \bm{A}_2 \hat{v}(\bm{x}) \bm{A}_2^T.
 \end{align}

\section{Additional Results}\label{app:additional_results}

\subsection{Comparison of Computation Times}

\begin{table}[h!]
\caption{Average compute time per BO iteration of different acquisition functions
as needed for the within-model comparison. We report the mean (std) across the
50 different function samples. All units are in seconds. Timing experiments
were run on an Intel Xeon CPU E5-1620 v4@3.50GHz.} \label{tab:compute_times}
\begin{center}
\begin{tabular}{ll}
Acquisition function                                   & time [sec] \\ \hline \\
NES-RS (ours)                                          & 5.39 (0.23) \\
NES-EP (ours)                                          & 1.90 (0.60) \\
BO-UU UCB \citep{Beland2017UncertaintyBONipsWorkshop}  & 0.06 (0.03) \\
BO-UU EI \citep{Beland2017UncertaintyBONipsWorkshop}   & 0.71 (0.33) \\
Unsc. BO \citep{Nogueira2016unscentedBO}               & 0.15 (0.09) \\
Standard BO EI                                         & 0.07 (0.03) \\
\end{tabular}
\end{center}
\end{table}

\subsection{Results for Hartmann (6-dim.)}

In \lfsecref{sec:results_synthetic_benchmark_functions} we provide a comparison on several benchmark functions up to three dimensions in terms of the inference regret, $r_n = | g(\textbf{x}_n^*) - g^* |$.
For computing the regret, one requires the 'true' robust optimum value g*. 
This value is generally not known and has to be found numerically.
In practice, we use the FFT over discrete signals to approximate the expectation in Equation~(1).
For the 3-dimensional Hartmann function, we use $n_{\text{FFT}} = 101$ evaluation points in each dimension to achieve high accuracy.
However, in 6 dimensions this is computationally infeasible for the required accuracy.
Thus, we compare the different acquisition functions just in terms of the estimated optimal robust value $g(\textbf{x}_n^*)$, see \lffigref{fig:convergence_results_hartmann6}.
The input noise was set to $\bm{\Sigma}_x = 0.1^2 \bm{I}$.

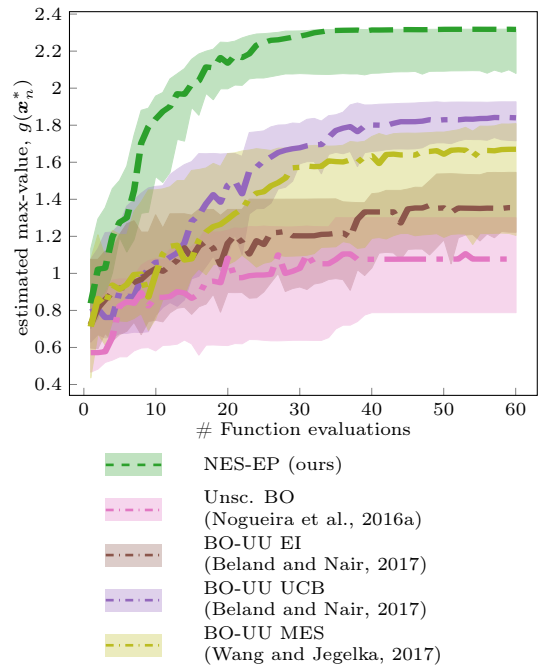
\begin{figure}
\centering
\begin{subfigure}[p]{.02\linewidth}
 {\figurefontsize\begin{tikzpicture}
\node[] at (0,0) {};
\node[rotate=90] at (0.0,1.0) {estimated max-value, $g(\bm{x}_n^*)$};
\end{tikzpicture}}
\end{subfigure}
\begin{subfigure}[p]{0.9\linewidth}
{\figurefontsize
\begin{tikzpicture}
\begin{axis}[
x label style={at={(0.5,-0.03)},anchor=south},
xmin=-2.0,xmax=63.,
xlabel=\# Function evaluations,
ylabel near ticks,
ymin=3.413e-1,ymax=2.413e-0,
major tick length=0.1cm,
minor tick length = 0.05cm,
tick pos=left,
height=\axisheight,
width=\axiswidth,
axis on top,
max space between ticks=20
]
\addplot[thick,blue] graphics[xmin=-2.0,xmax=63.,ymin=3.413e-1,ymax=2.413e-0] {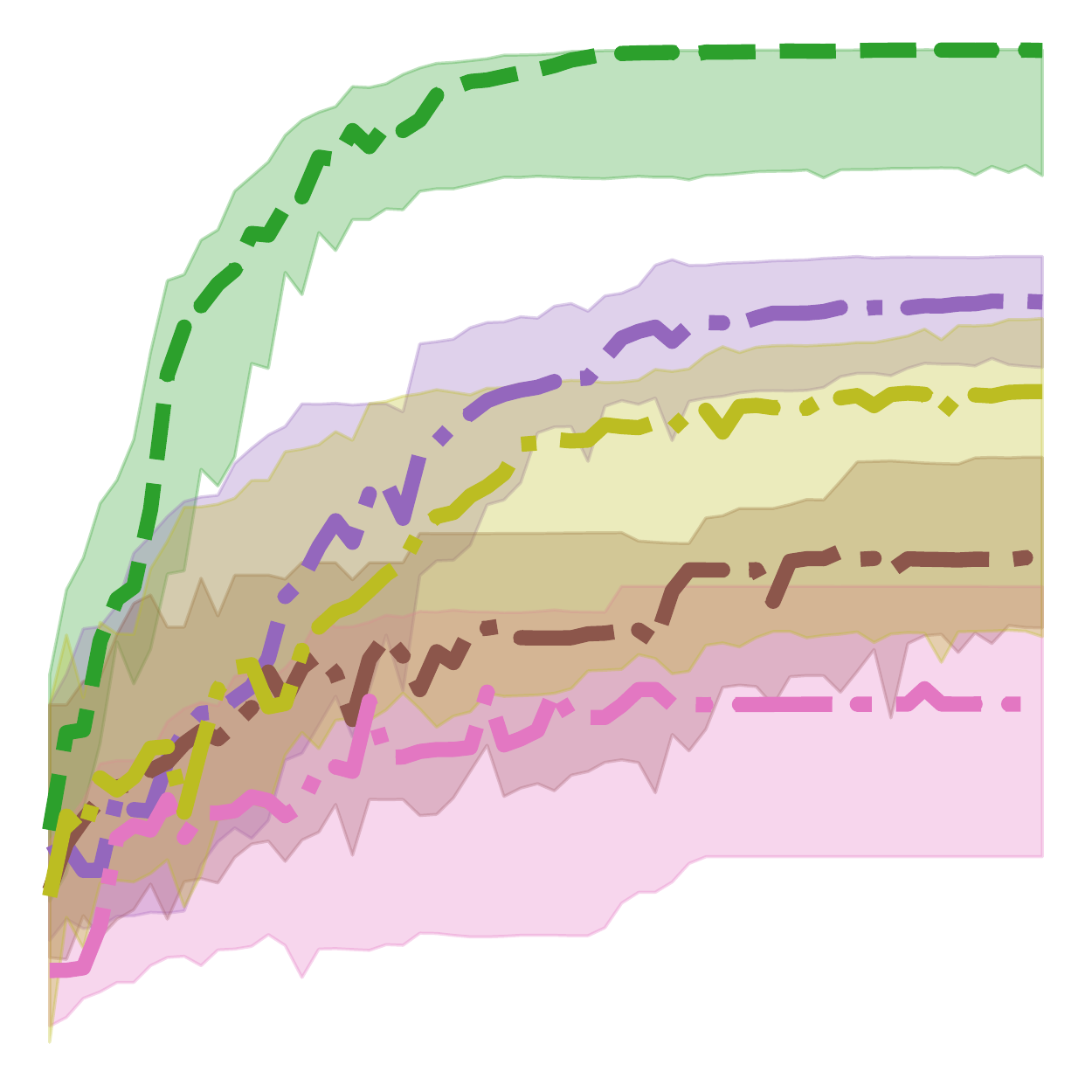};
\end{axis}
\end{tikzpicture}
}
\end{subfigure}
\begin{subfigure}[p]{\axiswidth}
 {\figurefontsize\begin{tikzpicture}

\def\barhalfheight{0.15}
\def\barwidth{0.75}
\def\dy{0.6}
\def\dxtext{2.7}
\def\dxbar{1.5}

\newcommand{\LegendList}{
 0/tableauC2/dashed/NES-EP (ours),
 1/tableauC6/dashdotted/Unsc. BO \\ \citep{Nogueira2016unscentedBO},
 2/tableauC5/dashdotted/BO-UU EI \\ \citep{Beland2017UncertaintyBONipsWorkshop},
 3/tableauC4/dashdotted/BO-UU UCB \\ \citep{Beland2017UncertaintyBONipsWorkshop},
 4/tableauC8/dashdotted/BO-UU MES \\ \citep{Wang2017MaxValueEntropySearch}
}

\node at (0.0, 0) {};
\foreach \i/\markercolor/\linestyle/\entry in \LegendList {
 \node[anchor=west, align=left] at (\dxtext,-\i*\dy) {\entry};
 \fill [\markercolor!30!white] (\dxbar,\barhalfheight-\i*\dy) rectangle (\dxbar + \barwidth, -\barhalfheight-\i*\dy);
 \draw [-,\markercolor, \linestyle, line width = 1.0pt] (\dxbar, -\i*\dy) -- (\dxbar + \barwidth, -\i*\dy);
}

\end{tikzpicture}}
\end{subfigure}
\caption{Estimated robust max-value $g(\bm{x}_n^*)$ for the 6-dimensional Hartmann function.
 We present the median (lines) and 25/75\textsuperscript{th} percentiles (shaded areas) across 20 independent runs with 10 randomly sampled initial points.}
\label{fig:convergence_results_hartmann6}
\end{figure}

\subsection{Number of Max-Value Samples}\label{app:num_maxvalue_samples}

\begin{figure}
\centering
\begin{subfigure}[p]{.02\linewidth}
 {\figurefontsize\begin{tikzpicture}
\node[] at (0,0) {};
\node[rotate=90] at (0.0,1.0) {inference regret};
\end{tikzpicture}}
\end{subfigure}
\begin{subfigure}[p]{0.9\linewidth}
{\figurefontsize
\begin{tikzpicture}
\begin{axis}[
x label style={at={(0.5,-0.03)},anchor=south},
xmin=-0.5,xmax=31.4,
xlabel=\# Function evaluations,
ymode = log,
ylabel near ticks,
ymin=5.356e-9,ymax=7.989e-1,
major tick length=0.1cm,
minor tick length = 0.05cm,
tick pos=left,
height=0.8\linewidth,
width=\axiswidth,
axis on top,
max space between ticks=20
]
\addplot[thick,blue] graphics[xmin=-0.5,xmax=31.4,ymin=5.356e-9,ymax=7.989e-1] {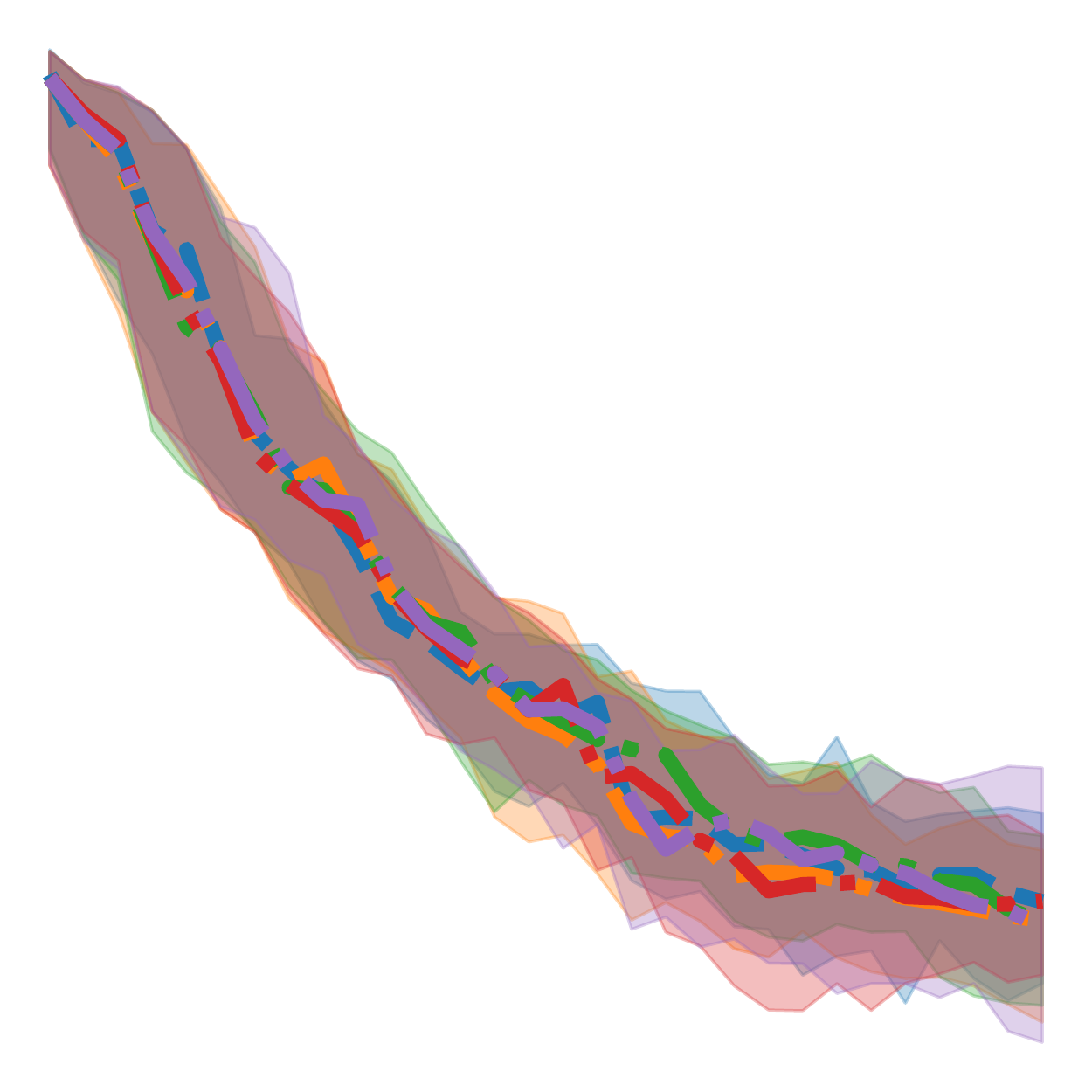};
\end{axis}
\end{tikzpicture}
}
\end{subfigure}
\begin{subfigure}[p]{\axiswidth}
 {\figurefontsize\begin{tikzpicture}

\def\barhalfheight{0.15}
\def\barwidth{0.75}
\def\dy{-0.5}
\def\dx{2.5}

\def\dxtext{1.5}
\def\dxbar{0.7}
\def\dxcols{3.5}

\newcommand{\LegendTopRow}{0/tableauC0/dashdotted/K=1,
                           1/tableauC1/dashdotted/K=3,
                           2/tableauC2/dashdotted/K=10}

\newcommand{\LegendBottomRow}{0/tableauC3/dashdotted/K=30,
                              1/tableauC4/dashdotted/K=100}

\node at (0, 0) {};
\foreach \i/\markercolor/\linestyle/\entry in \LegendTopRow {
 \node[anchor=west] at (\dxtext + \i*\dx ,0.0) {\entry};
 \fill [\markercolor!30!white] (\dxbar + \i*\dx,\barhalfheight) rectangle (\dxbar + \barwidth + \i*\dx, -\barhalfheight);
 \draw [-,\markercolor, \linestyle, line width = 1.0pt] (\dxbar + \i*\dx, 0.0) -- (\dxbar + \barwidth + \i*\dx, 0.0);
}

\foreach \i/\markercolor/\linestyle/\entry in \LegendBottomRow {
 \node[anchor=west] at (\dxtext + \i*\dx , \dy) {\entry};
 \fill [\markercolor!30!white] (\dxbar + \i*\dx,\barhalfheight + \dy) rectangle (\dxbar + \barwidth + \i*\dx, -\barhalfheight  + \dy);
 \draw [-,\markercolor, \linestyle, line width = 1.0pt] (\dxbar + \i*\dx, \dy) -- (\dxbar + \barwidth + \i*\dx, \dy);
}

\end{tikzpicture}}
\end{subfigure}
\caption{Within-model comparison in terms of the inference regret $r_n = |g(\bm{x}_n^*) - g^*|$ for different values of the hyperparameter $K$, i.e., the number of Monte-Carlo samples to approximate the expectation over robust max-values.
 As there is no significant difference in the performance, we used $K=1$ for all experiments in the paper due to the lower computational cost.}
 \label{fig:appendix_nes_ep_comparison}
\end{figure}
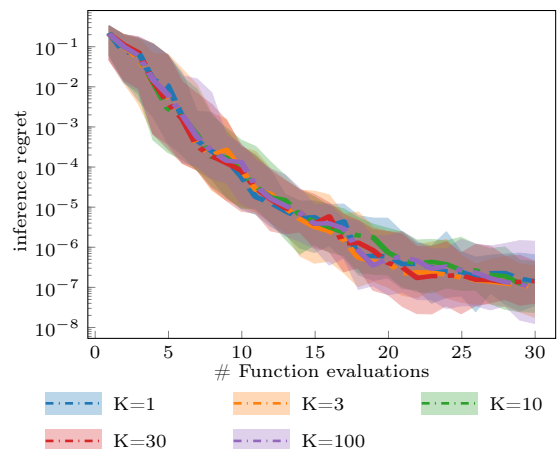
In Section~3.1 we discuss how to approximate the expectation over robust maximum values by Monte Carlo sampling.
Here, we explain the exact sampling procedure and subsequently present results of a within-model comparison that investigates the effect of the number of robust max-value samples $K$ on the final result.

\paragraph{Sampling Max-Values}
Note that the computation of the acquisition function scales linearly with the number $K$ of Monte-Carlo samples.
However, sampling the robust max-values only needs to be done once per \gls{acr:bo} iteration, while the acquisition function requires many evaluations during one \gls{acr:bo} iteration.
Thus, it is advantageous to use as few Monte-Carlo samples as possible.
The exact sampling procedure for $K$ robust max-value samples is given as follows:
\begin{enumerate}
 \item Sample $100$ robust max-values as described in Section~3.1,
 \item Create a regular grid between the 25\textsuperscript{th} and 75\textsuperscript{th} percentile with $K$ points,
 \item Draw the robust max-values from the sample distribution (step 1) corresponding to the percentiles of the regular grid (step 2).
\end{enumerate}
The benefit of this procedure is that it makes the estimate of the expectation more robust w.r.t. the number of samples used.

\paragraph{Within-Model Comparison}
To investigate the effect of the number of Monte-Carlo samples $K$ on the final performance, we perform a within-model comparison for NES-EP with $K = \{1, 3, 10, 30, 100\}$ samples.
Results are presented in \lffigref{fig:appendix_nes_ep_comparison}.
Note that the performance is independent of the number of samples used to approximate the expectation.
Thus, for the purpose of computational efficiency we use $K=1$ for all experiments in the paper.

\subsection{Unscented BO: Hyperparameter $\bm{\kappa}$}
The unscented transformation \citep{Julier2004UnscentedFiltering} used for unscented \gls{acr:bo} \citep{Nogueira2016unscentedBO} is based on a weighted sum:
\begin{align}
 \bar{\bm{x}} = \mathbb{E}_{\bm{x}} \left[ f(\bm{x}) \right] \approx \sum_{i=0}^{2d} \omega^{(i)} f(\bm{x}^{(i)}),
\end{align}
with $\bm{x} \sim \mathcal{N}(\bm{x} | \bm{x}^0, \bm{\Sigma}_x)$.
The so-called sigma points $\bm{x}^{(i)}$ are computed as
\begin{align}
\begin{split}
 \bm{x}_{+}^{(i)} & = \bm{x}^0 + \left( \sqrt{(d+\kappa) \bm{\Sigma}_x} \right)_i, \quad \forall i = 1, \dots, d \\ 
 \bm{x}_{-}^{(i)} & = \bm{x}^0 - \left( \sqrt{(d+\kappa) \bm{\Sigma}_x} \right)_i, \quad \forall i = 1, \dots, d,
\end{split}
\end{align}
where $(\sqrt{\cdot})_i$ is the $i$-th column of the (elementwise) square root of the corresponding matrix.
The weights $\omega^{(i)}$ to the corresponding sigma points are given by
\begin{align}\label{eq:sigma_weights}
\begin{split}
 \omega^0 & = \frac{k}{d + \kappa}, \\
 \omega_{+}^{(i)} =  \omega_{-}^{(i)} & = \frac{1}{2(d + \kappa)}, \quad \forall i = 1, \dots, d.
\end{split}
\end{align}
In the corresponding tech-report \citep{Nogueira2016unscentedBOArxiv} to the original paper \citep{Nogueira2016unscentedBO}, the authors discuss the choice of optimal values for the hyperparameter $k$ and suggest $\kappa=0.0$ or $\kappa=-3.0$.
For negative (integer) values of $k$, however, \lfeqref{eq:sigma_weights} leads to a division by zero if $d = -\kappa$.
Thus, we decided against $\kappa=-3.0$ to be consistent across all experiments and objective functions.
To find the best (non-negative) value for $\kappa$ we performed a within-model comparison with different values for $\kappa$ in the range between 0.0 and 2.0.
Results are presented in \lffigref{fig:appendix_unsc_bo_comparison}.
We found that for $\kappa=1.0$, unscented \gls{acr:bo} showed the best performance and consequently also used $\kappa=1.0$ for all experiments in the paper.

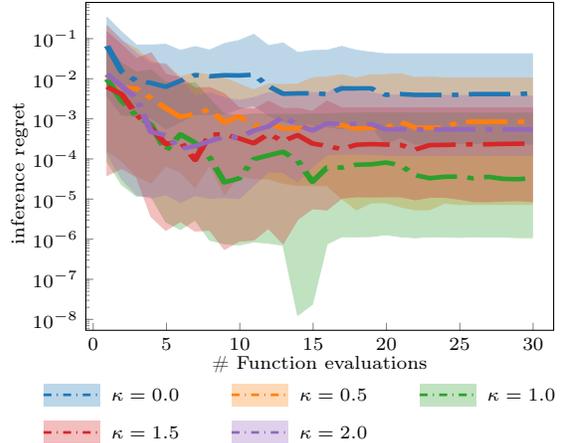
\begin{figure}
\centering
\begin{subfigure}[p]{.02\linewidth}
 {\figurefontsize\begin{tikzpicture}
\node[] at (0,0) {};
\node[rotate=90] at (0.0,1.0) {inference regret};
\end{tikzpicture}}
\end{subfigure}
\begin{subfigure}[p]{0.9\linewidth}
{\figurefontsize
\begin{tikzpicture}
\begin{axis}[
x label style={at={(0.5,-0.03)},anchor=south},
xmin=-0.5,xmax=31.4,
xlabel=\# Function evaluations,
ymode = log,
ylabel near ticks,
ymin=5.356e-9,ymax=7.989e-1,
major tick length=0.1cm,
minor tick length = 0.05cm,
tick pos=left,
height=0.8\linewidth,
width=\axiswidth,
axis on top,
max space between ticks=20
]
\addplot[thick,blue] graphics[xmin=-0.5,xmax=31.4,ymin=5.356e-9,ymax=7.989e-1] {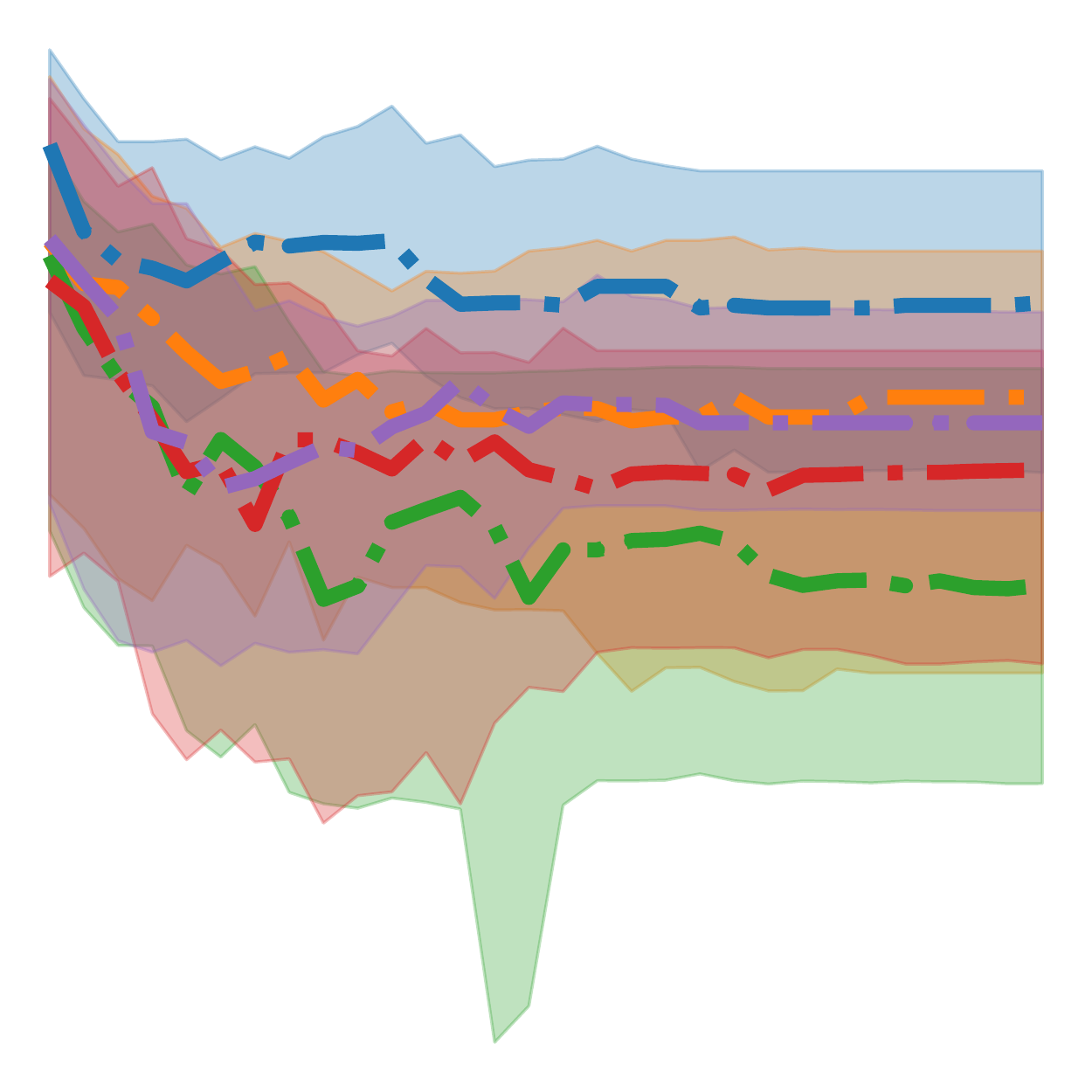};
\end{axis}
\end{tikzpicture}
}
\end{subfigure}
\begin{subfigure}[p]{\axiswidth}
 {\figurefontsize\begin{tikzpicture}

\def\barhalfheight{0.15}
\def\barwidth{0.75}
\def\dy{-0.5}
\def\dx{2.5}

\def\dxtext{1.5}
\def\dxbar{0.7}
\def\dxcols{3.5}

\newcommand{\LegendTopRow}{0/tableauC0/dashdotted/\kappa=0.0,
 1/tableauC1/dashdotted/\kappa=0.5,
 2/tableauC2/dashdotted/\kappa=1.0}

\newcommand{\LegendBottomRow}{0/tableauC3/dashdotted/\kappa=1.5,
 1/tableauC4/dashdotted/\kappa=2.0}

\node at (0, 0) {};
\foreach \i/\markercolor/\linestyle/\entry in \LegendTopRow {
 \node[anchor=west] at (\dxtext + \i*\dx ,0.0) {$\entry$};
 \fill [\markercolor!30!white] (\dxbar + \i*\dx,\barhalfheight) rectangle (\dxbar + \barwidth + \i*\dx, -\barhalfheight);
 \draw [-,\markercolor, \linestyle, line width = 1.0pt] (\dxbar + \i*\dx, 0.0) -- (\dxbar + \barwidth + \i*\dx, 0.0);
}

\foreach \i/\markercolor/\linestyle/\entry in \LegendBottomRow {
 \node[anchor=west] at (\dxtext + \i*\dx , \dy) {$\entry$};
 \fill [\markercolor!30!white] (\dxbar + \i*\dx,\barhalfheight + \dy) rectangle (\dxbar + \barwidth + \i*\dx, -\barhalfheight  + \dy);
 \draw [-,\markercolor, \linestyle, line width = 1.0pt] (\dxbar + \i*\dx, \dy) -- (\dxbar + \barwidth + \i*\dx, \dy);
}

\end{tikzpicture}}
\end{subfigure}
\caption{Within-model comparison in terms of the inference regret $r_n = |g(\bm{x}_n^*) - g^*|$ for different values of the hyperparameter $K$, i.e., the number of Monte-Carlo samples to approximate the expectation over robust max-values.
 As there is no significant difference in the performance, we used $K=1$ for all experiments in the paper due to the lower computational cost.}
 \label{fig:appendix_unsc_bo_comparison}
\end{figure}

\subsection{Synthetic Benchmark Functions - Distance to Robust Optimum}
In the main part of this paper, we compare all methods with respect to the inference regret  \mbox{$r_n = | g(\bm{x}_n^*) - g^* |$}.
Depending on the objective's scale, the inference regret may be small although an entirely different optimum is found.
Here, we present the results in terms of distance to the optimum $\norm{\bm{x}_n^* - \bm{x}^*}.$
See Sec.~4.1 for details on the objective functions and the evaluated methods.

\begin{figure*}[h!]
\centering
\begin{subfigure}[p]{.02\linewidth}
 {\figurefontsize\begin{tikzpicture}
\node[] at (0,0) {};
\node[rotate=90] at (0.0,1.5) {distance to optimum};
\end{tikzpicture}}
\end{subfigure}
\begin{subfigure}[p]{\subfigurewidth}
{\figurefontsize
\begin{tikzpicture}
\begin{axis}[
x label style={at={(0.5,-0.02)},anchor=south},
xmin=0.0,xmax=20.9,
xlabel=\# Function evaluations,
ymode = log,
ylabel near ticks,
ymin=1.286e-5,ymax=1.084e-0,
major tick length=0.1cm,
minor tick length = 0.05cm,
tick pos=left,
height=\axisheight,
width=\axiswidth,
axis on top,
max space between ticks=20
]
\addplot[thick,blue] graphics[xmin=0.0,xmax=20.9,ymin=1.286e-5,ymax=1.084e-0,] {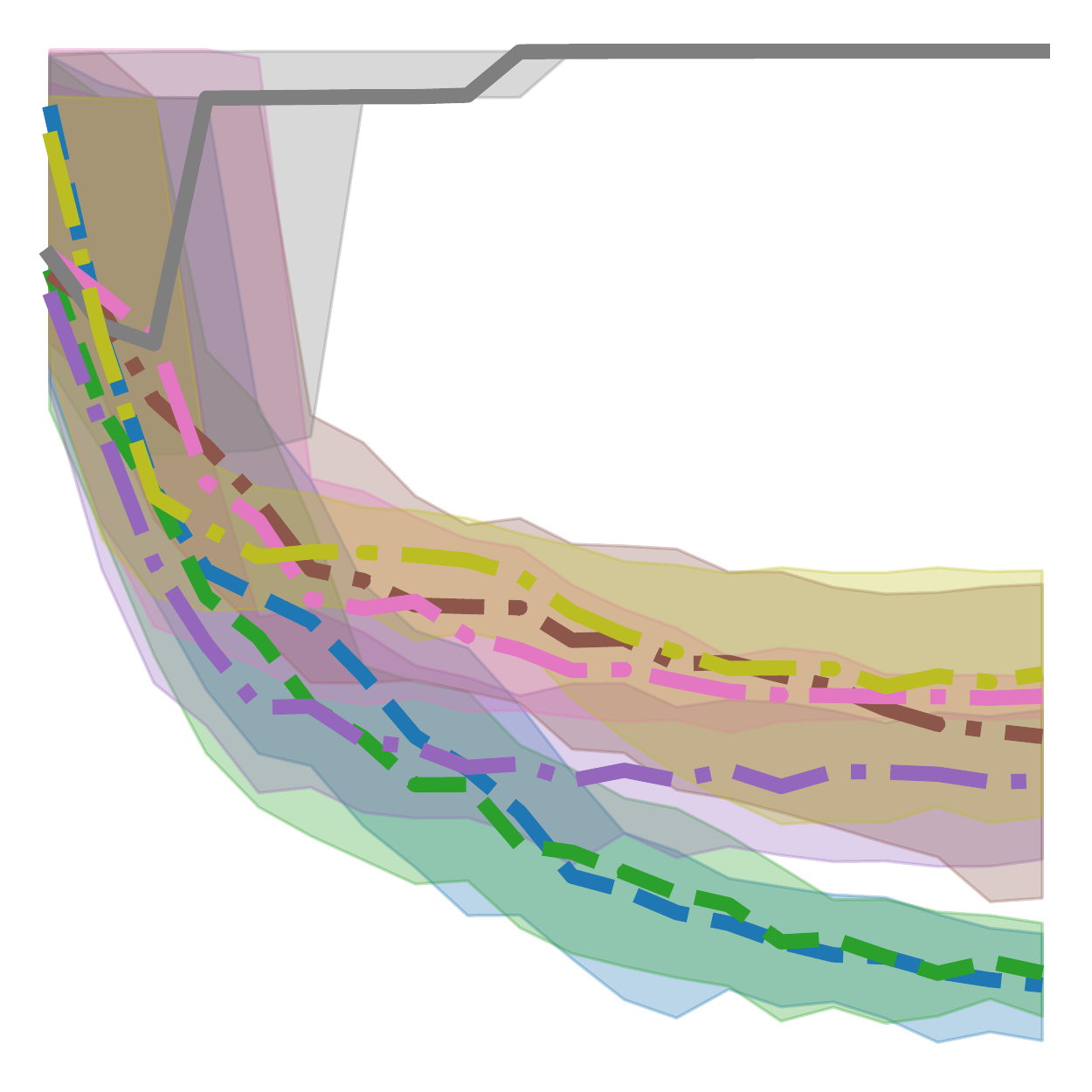};
\end{axis}
\end{tikzpicture}
}
 \caption{Sin + Linear (1-dim.)}\label{fig:dx_results_convergence_synthetic_1d_01}
\end{subfigure}
\begin{subfigure}[p]{\subfigurewidth}
{\figurefontsize
 \begin{tikzpicture}
 \begin{axis}[
 x label style={at={(0.5,-0.02)},anchor=south},
 xmin=-0.5,xmax=31.4,
 xlabel=\# Function evaluations,
 ymode = log,
 ylabel near ticks,
 ymin=4.923e-6,ymax=8.767e-0,
 major tick length=0.1cm,
 minor tick length = 0.05cm,
 tick pos=left,
 height=\axisheight,
 width=\axiswidth,
 axis on top,
 max space between ticks=20
 ]
 \addplot[thick,blue] graphics[xmin=-0.5,xmax=31.4,ymin=4.923e-6,ymax=8.767e-0,] {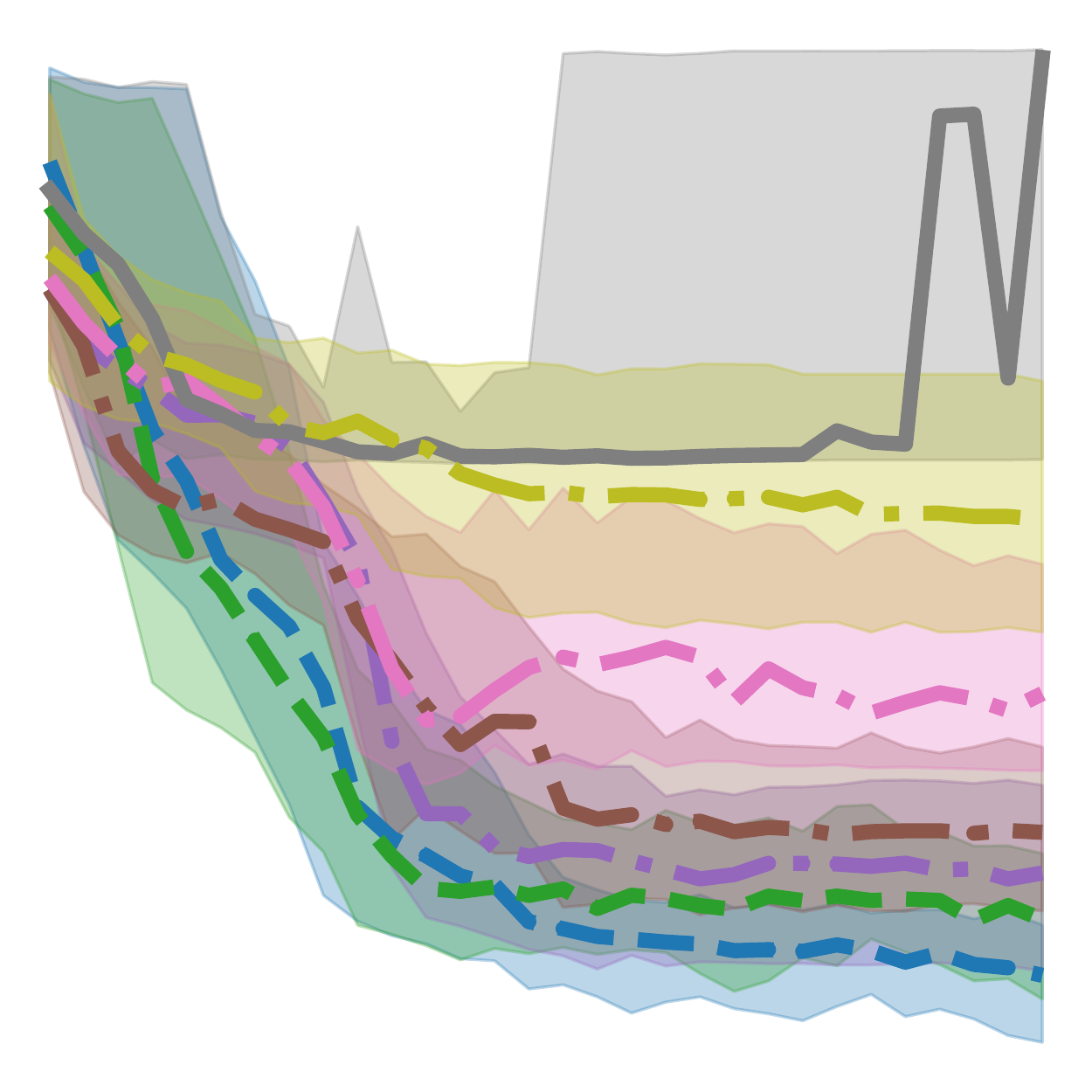};
 \end{axis}
 \end{tikzpicture}
}
 \caption{RKHS-function (1-dim.)}\label{fig:dx_results_convergence_rkhs_synth}
\end{subfigure}
\begin{subfigure}[p]{\subfigurewidth}
{\figurefontsize
 \begin{tikzpicture}
 \begin{axis}[
 x label style={at={(0.5,-0.02)},anchor=south},
 xmin=-0.5,xmax=31.4,
 xlabel=\# Function evaluations,
 ymode = log,
 ylabel near ticks,
  ymin=6.581e-4,ymax=8.288e-1,
 major tick length=0.1cm,
 minor tick length = 0.05cm,
 tick pos=left,
 height=\axisheight,
 width=\axiswidth,
 axis on top,
 max space between ticks=20
 ]
 \addplot[thick,blue] graphics[xmin=-0.5,xmax=31.4,ymin=6.581e-4,ymax=8.288e-1,] {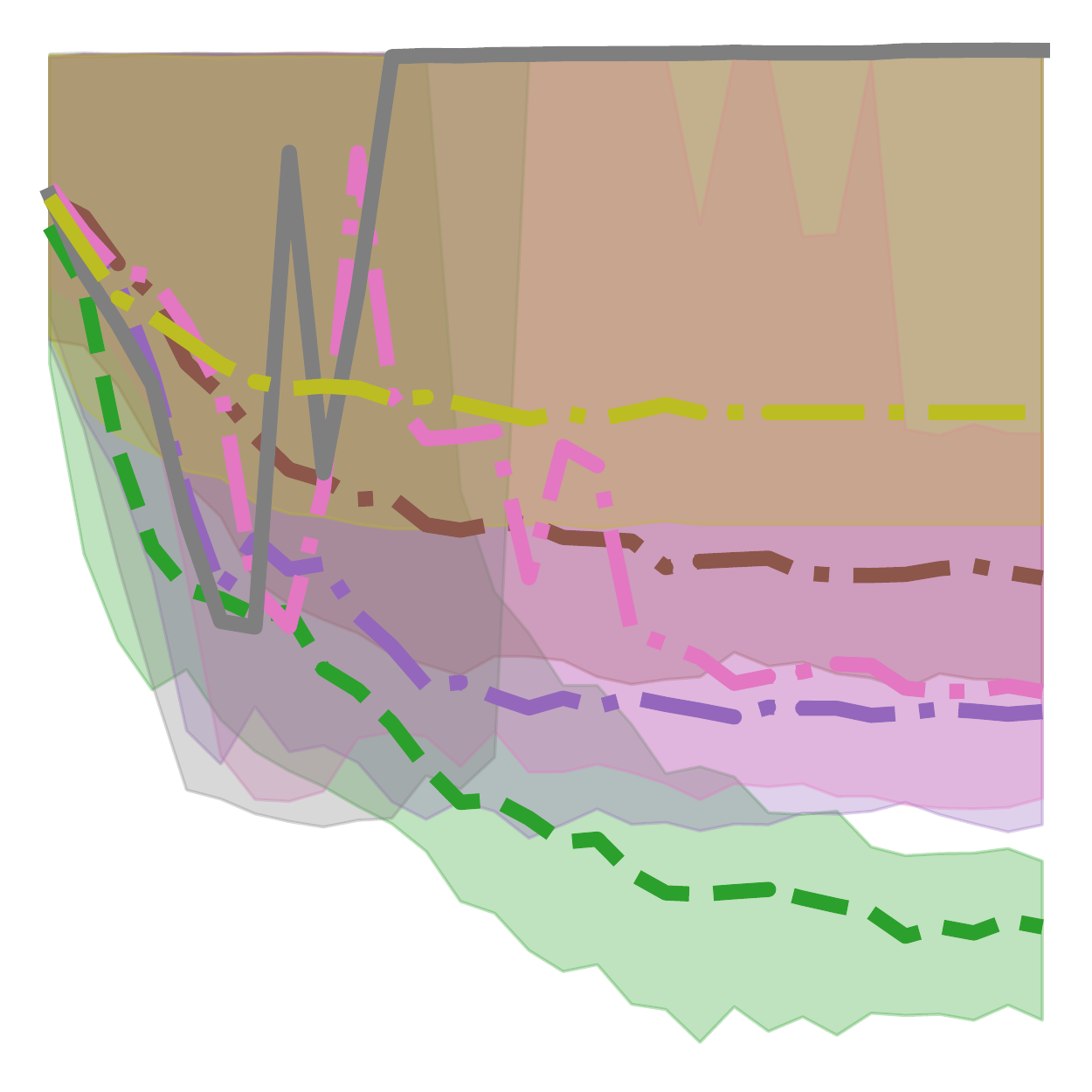};
 \end{axis}
 \end{tikzpicture}
}
 \caption{GMM (2-dim.)}\label{fig:dx_results_convergence_gmm_2d}
\end{subfigure}

\vspace*{5mm}

\begin{subfigure}[p]{.02\linewidth}
 {\figurefontsize\begin{tikzpicture}
\node[] at (0,0) {};
\node[rotate=90] at (0.0,1.5) {distance to optimum};
\end{tikzpicture}}
\end{subfigure}
\begin{subfigure}[p]{\subfigurewidth}
 {\figurefontsize
  \begin{tikzpicture}
  \begin{axis}[
  x label style={at={(0.5,-0.02)},anchor=south},
  xlabel near ticks, 
  xmin=-1.5,xmax=52.5,
  xlabel=\# Function evaluations,
  ymode = log,
  ylabel near ticks,
  ymin=4.866e-2,ymax=5.162e+0,
  major tick length=0.1cm,
  minor tick length = 0.05cm,
  tick pos=left,
  height=\axisheight,
  width=\axiswidth,
  axis on top,
  max space between ticks=20
  ]
  \addplot[thick,blue] graphics[xmin=-1.5,xmax=52.5,ymin=4.866e-2,ymax=5.162e+0,] {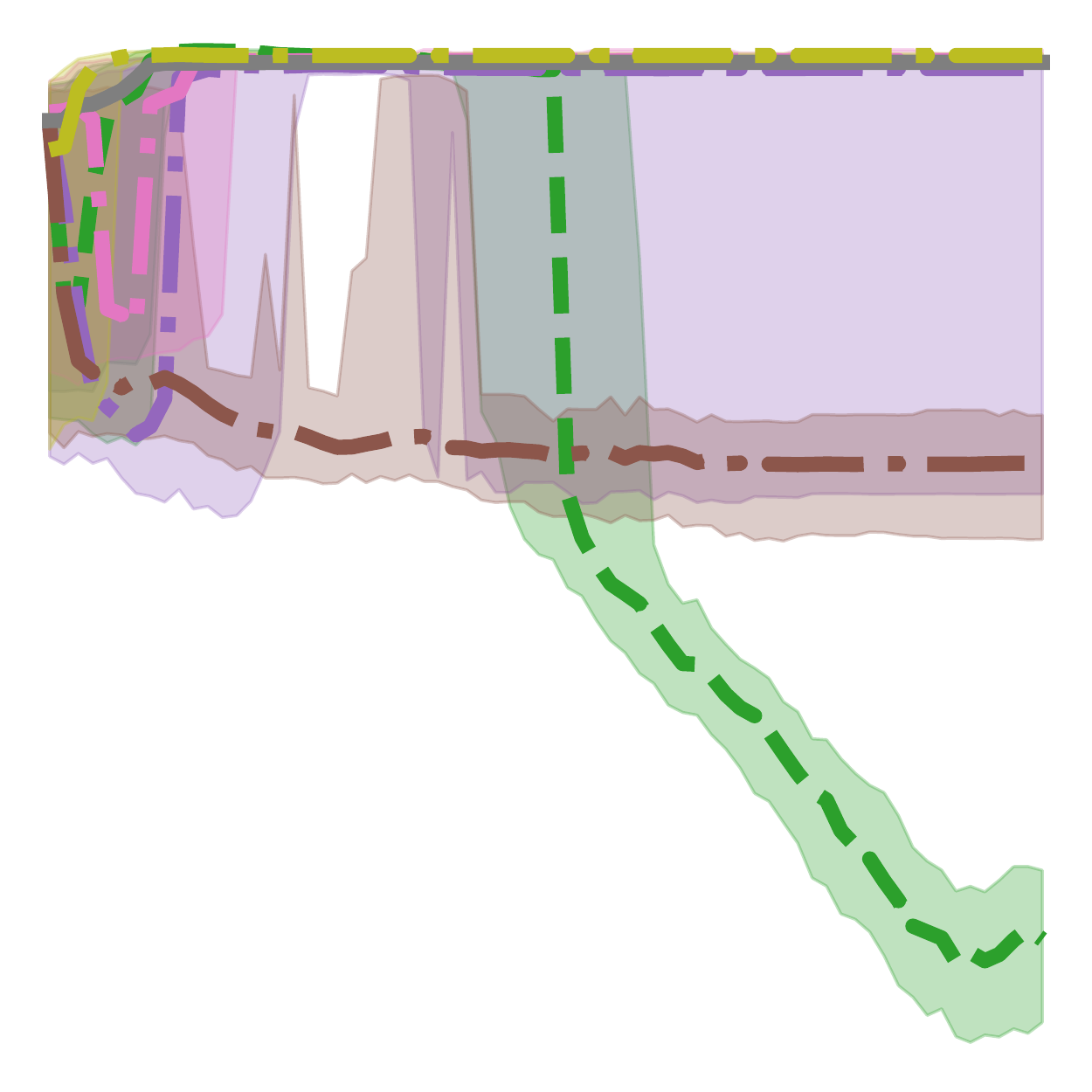};
  \end{axis}
  \end{tikzpicture}
 }
\vspace*{\captionvspace}
 \caption{Polynomial (2-dim.)}\label{fig:dx_results_convergence_synth_poly_2d}
\end{subfigure}
\begin{subfigure}[p]{\subfigurewidth}
 {\figurefontsize
  \begin{tikzpicture}
  \begin{axis}[
  x label style={at={(0.5,-0.02)},anchor=south},
  xlabel near ticks, 
  xmin=-4.0,xmax=105.0,
  xlabel=\# Function evaluations,
  ymode = log,
  ylabel near ticks,
       ytick={1e-3, 1e-2, 1e-1},
  ymin=9.099e-4,ymax=9.829e-1,
  major tick length=0.1cm,
  minor tick length = 0.05cm,
  tick pos=left,
  height=\axisheight,
  width=\axiswidth,
  axis on top,
  max space between ticks=20
  ]
  \addplot[thick,blue] graphics[xmin=-4.0,xmax=105.0,ymin=9.099e-4,ymax=9.829e-1,] {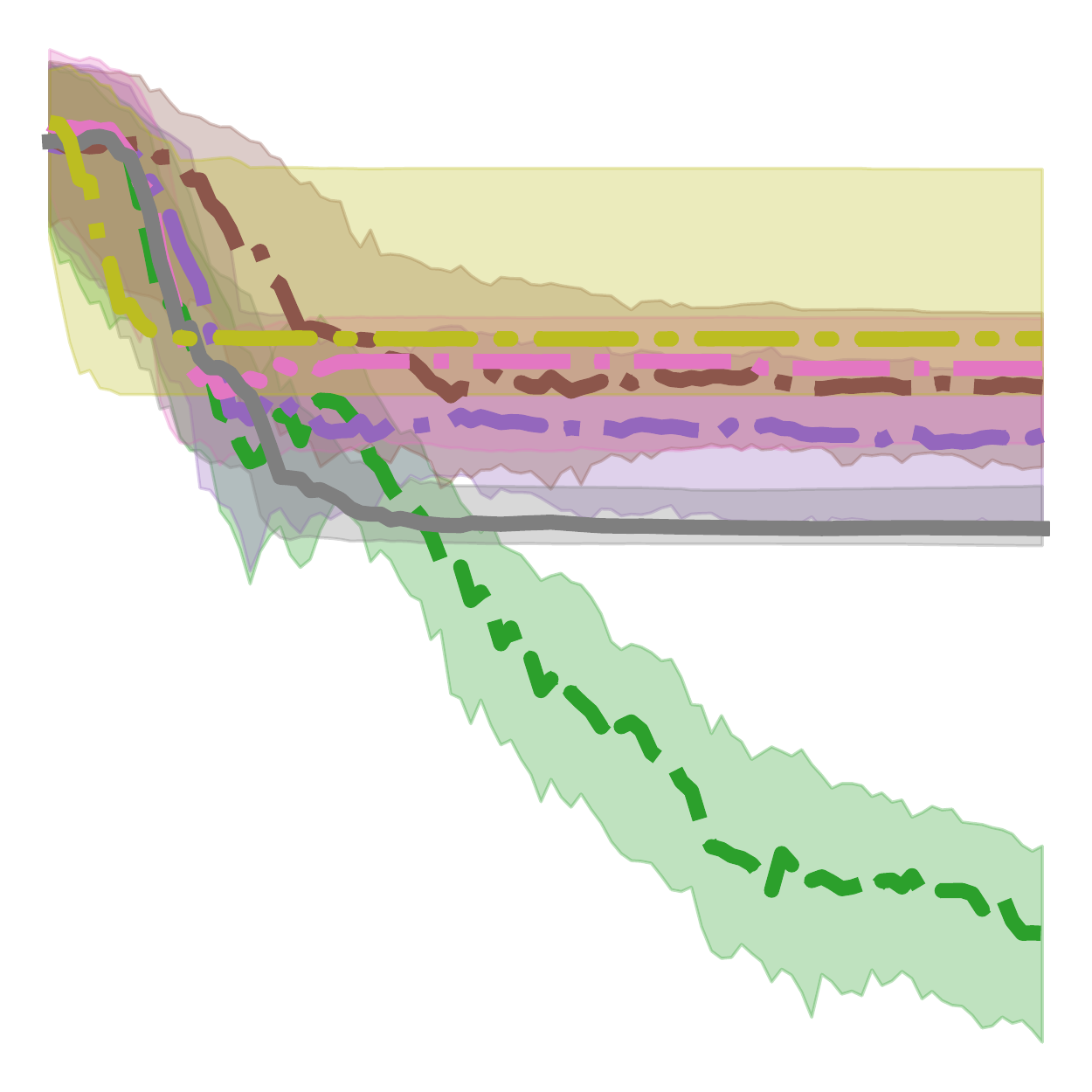};
  \end{axis}
  \end{tikzpicture}
 }
\vspace*{\captionvspace}
 \caption{Hartmann (3-dim.)}\label{fig:dx_results_convergence_hartmann_3d}
\end{subfigure}
\begin{subfigure}[p]{\subfigurewidth}
 \vspace*{-0.0cm}
 \hspace*{+0.5cm}
 {\figurefontsize\begin{tikzpicture}

\def\barhalfheight{0.15}
\def\barwidth{0.75}
\def\dy{0.6}
\def\dxtext{1.2}
\def\dxbar{0.0}

\newcommand{\LegendList}{0/tableauC0/dashed/NES-RS (ours), 
                         1/tableauC2/dashed/NES-EP (ours),
                         2/tableauC6/dashdotted/Unscented BO \\ \citep{Nogueira2016unscentedBO},
                         3/tableauC5/dashdotted/BO-UU EI \\ \citep{Beland2017UncertaintyBONipsWorkshop},
                         4/tableauC4/dashdotted/BO-UU UCB \\ \citep{Beland2017UncertaintyBONipsWorkshop},
                         5/tableauC8/dashdotted/BO-UU MES \\
                         \citep{Wang2017MaxValueEntropySearch},
                         6/tableauC7/solid/Standard BO EI}

\node at (0.0, 0) {};
\foreach \i/\markercolor/\linestyle/\entry in \LegendList {
 \node[anchor=west, align=left] at (\dxtext,-\i*\dy) {\entry};
 \fill [\markercolor!30!white] (\dxbar,\barhalfheight-\i*\dy) rectangle (\dxbar + \barwidth, -\barhalfheight-\i*\dy);
 \draw [-,\markercolor, \linestyle, line width = 1.0pt] (\dxbar, -\i*\dy) -- (\dxbar + \barwidth, -\i*\dy);
}
\node at (0, -3.5) {};

\end{tikzpicture}}
\end{subfigure}
\caption{Distance to optimum $\norm{ \bm{x}_n^* - \bm{x}^*}_2$ on synthetic benchmark problems.
 We present the median (lines) and 25/75\textsuperscript{th} percentiles (shaded areas) across 100 independent runs with randomly sampled initial points.
}
\label{fig:dx_results_synthetic_functions_convergence}
\end{figure*}

\section{Synthetic Objective Functions}\label{app:synthetic_objective_functions}
In this section, the 1- and 2-dimensional functions $f(\bm{x})$ of the synthetic benchmark problems are visualized.
Furthermore, the robust counterparts $g(\bm{x})$ are depicted.
\begin{enumerate}[(a),leftmargin=0.7cm]
 \item $f(\bm{x}) = \sin(5 \pi \bm{x}^2) + 0.5 \bm{x}$, with $\bm{x} \in [0, 1]$ and $\bm{\Sigma}_x = 0.05^2$,
 \item RKHS-function (1-dim.) with $\bm{\Sigma}_x = 0.03^2$ from \cite{Assael2014HeteroscedasticBO}, also used by \citet{Nogueira2016unscentedBO},
 \item Gaussian mixture model (2-dim.)  with $\bm{\Sigma}_x = 0.1^2 \bm{I}$, also used by \citet{Nogueira2016unscentedBO},
 \item Polynomial (2-dim.) with $\bm{\Sigma}_x = 0.6^2 \bm{I}$ from \cite{Bertsimas2010RobustOptimization}, also used by \citet{Bogunovic2018AdversariallyBO}.
 We chose the domain to be $\mathcal{X} = [-0.75, -0.25] \times [3.0, 4.2]$ and scaled/shifted the original objective $f(x)$ s.t. $\mathbb{E}[f(x)] = 0.0$ and $\mathbb{V}[f(x)] = 1.0$.
\end{enumerate}
\begin{figure*}[h!]
 \centering
 \begin{subfigure}[p]{.35\linewidth}
  \includegraphics[width=\linewidth]{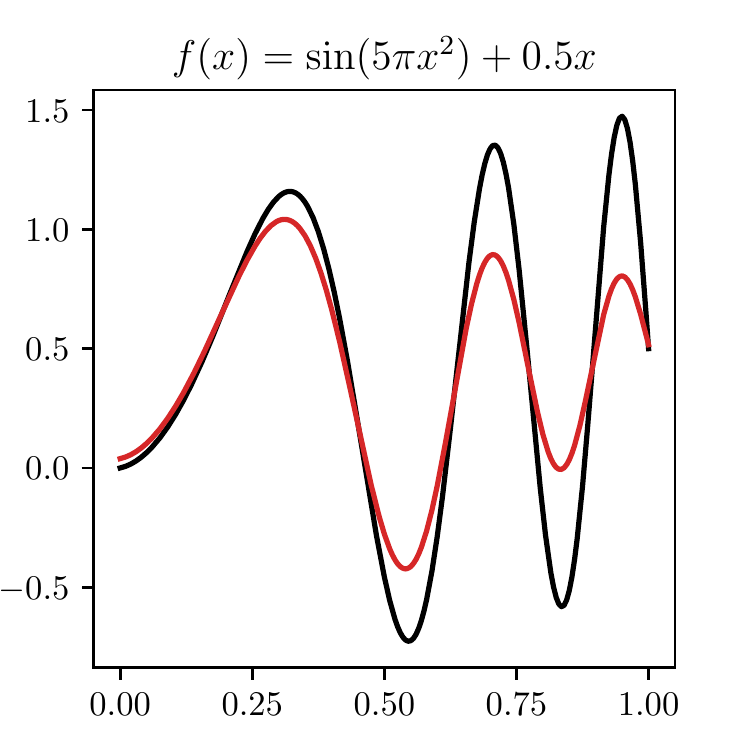}
  \caption{Sin + Linear (1-dim.). Black: synthetic function $f(\bm{x})$, red: robust counterpart $g(\bm{x})$.}\label{fig:visualization_synthetic_1d_01}
 \end{subfigure}\hspace*{1cm}
 \begin{subfigure}[p]{.35\linewidth}
  \includegraphics[width=\linewidth]{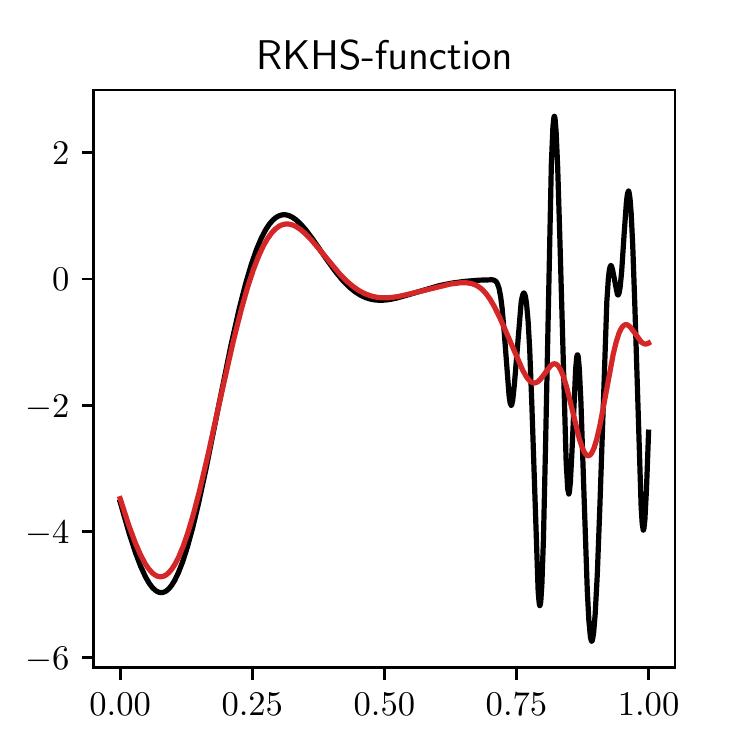}
  \caption{RKHS-function (1-dim.). Black: synthetic function $f(\bm{x})$, red: robust counterpart $g(\bm{x})$.}\label{fig:visualization_rkhs_synth}
 \end{subfigure}
 
 \begin{subfigure}[p]{.65\linewidth}
  \includegraphics[width=\linewidth]{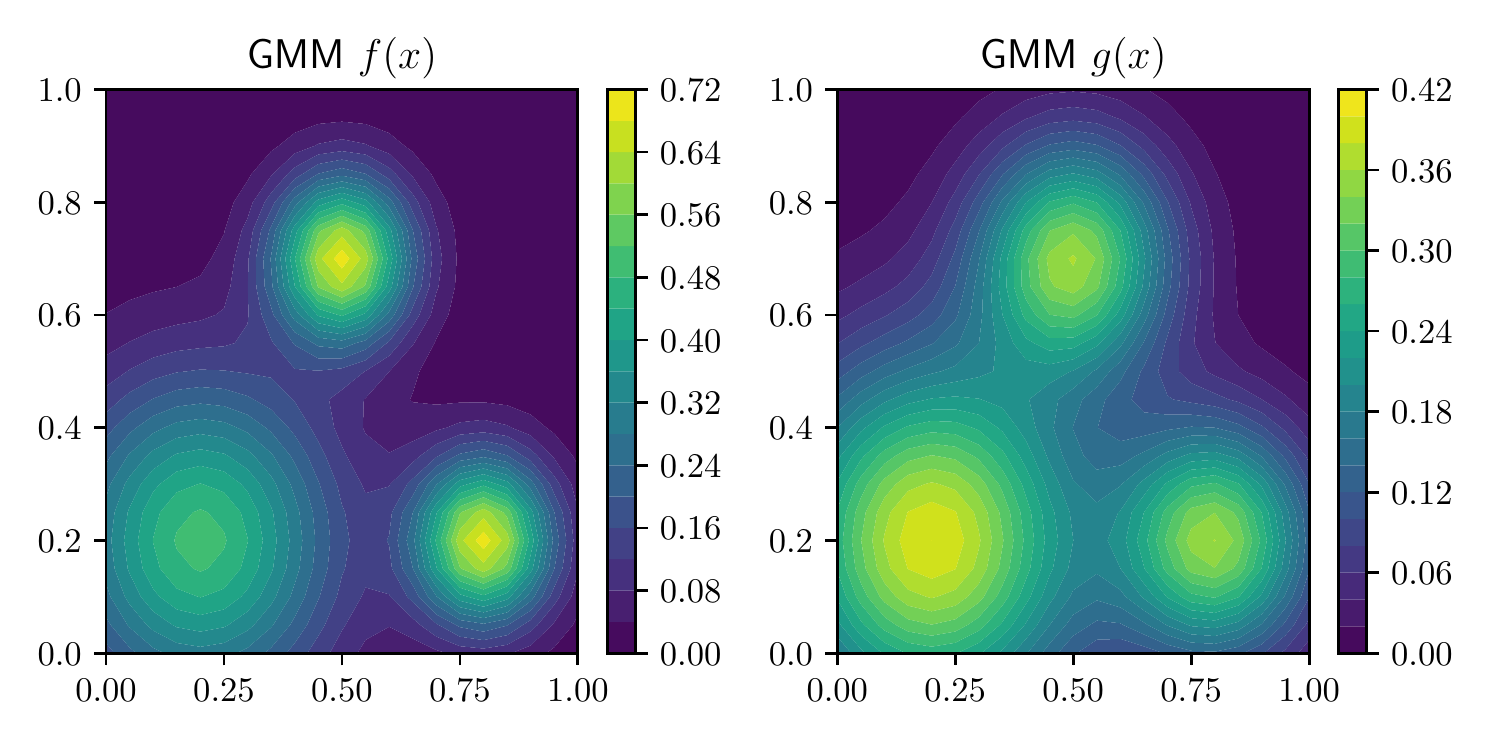}
  \caption{Gaussian Mixture Model (GMM) (2-dim.). Left: synthetic function $f(\bm{x})$, right: robust counterpart $g(\bm{x})$.}\label{fig:visualization_gmm_2d}
 \end{subfigure}
 
 \begin{subfigure}[p]{.65\linewidth}
  \includegraphics[width=\linewidth]{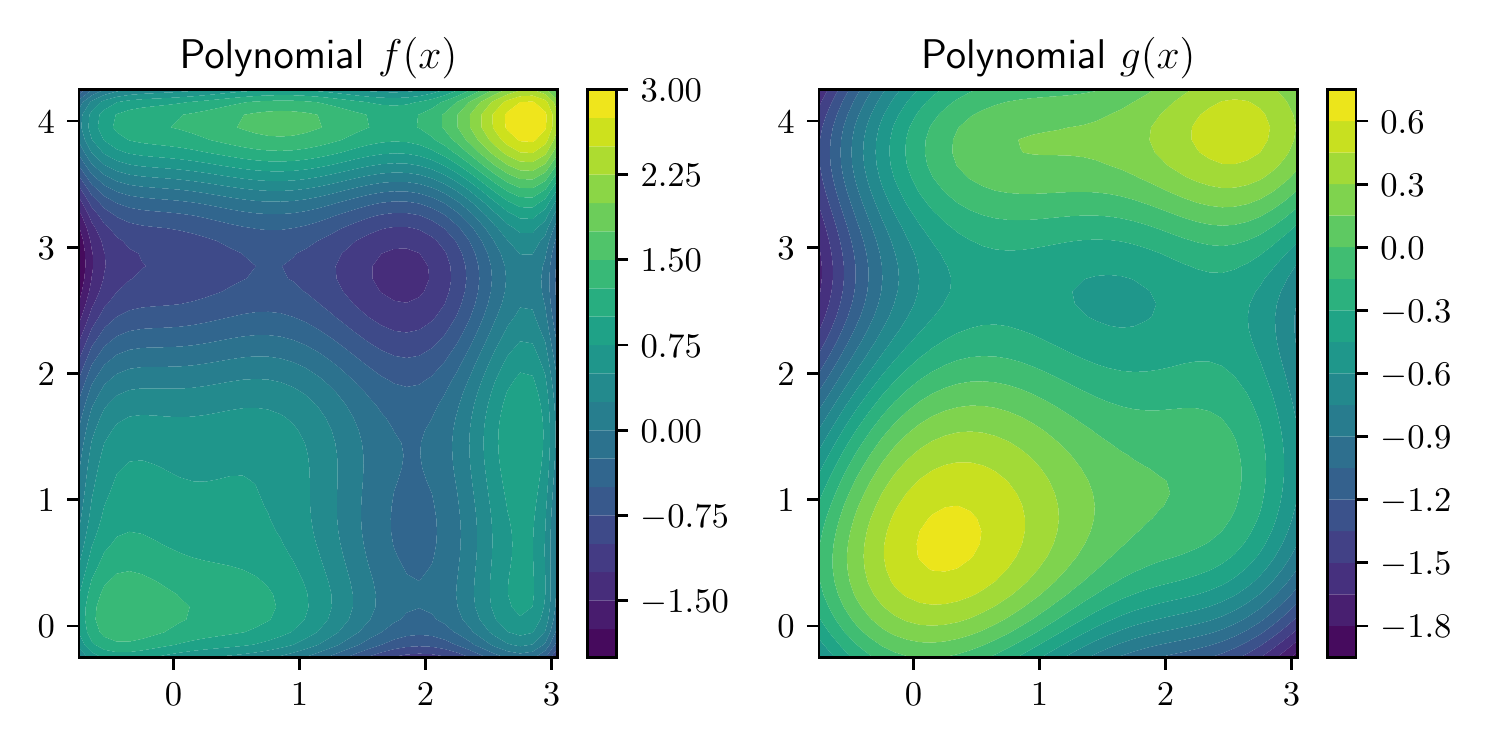}
  \caption{Polynomial (2-dim.). Left: synthetic function $f(\bm{x})$, right: robust counterpart $g(\bm{x})$.}\label{fig:visualization_synth_poly_2d}
 \end{subfigure}
 \caption{Visualization of synthetic benchmark functions $f(\bm{x})$ with the robust counterpart $g(\bm{x})$.}
  \label{fig:visualization_benchmarks}
\end{figure*}
}{}

\end{document}